\documentclass[twoside,11pt]{article}

\usepackage[arxiv]{melba}

\usepackage{amssymb,amsmath,amsfonts}
\usepackage{multirow}
\usepackage{subcaption}
\usepackage{pgfplots}
    \pgfplotsset{
        compat=newest
    }
\usepackage{tikz}
\usepackage{booktabs}
\usepackage{makecell}

\DeclareMathOperator*{\argmin}{arg\,min}

\melbaid{YYYY:NNN}  
\doi{https://doi.org/10.59275/j.melba.2023-AAAA}
\melbaauthors{Mounsaveng and Laradji and Vázquez and Perdersoli and Ben Ayed}  
\volume{2}
\firstpageno{0001}  
\melbayear{2023}  
\datesubmitted{07/2023}  
\datepublished{m2/yyyy}  

\ShortHeadings{Automatic DA Learning using Bilevel Optimization for Histopathological Images}{Mounsaveng and Laradji and Vázquez and Perdersoli and Ben Ayed}

\title{Automatic Data Augmentation Learning using \\Bilevel Optimization for Histopathological Images}

\author{\firstname Saypraseuth \surname Mounsaveng \email saypraseuth.mounsaveng.1@etsmtl.net \\  
	\addr ÉTS Montréal, Canada
	\AND
	\firstname Issam \surname Laradji \email issam.laradji@servicenow.com \\
	\addr ServiceNow Research
        \AND 
	\firstname David \surname Vázquez \email david.vazquez@servicenow.com \\
	\addr ServiceNow Research
 	\AND
        \firstname Marco \surname Pedersoli \email marco.pedersoli@etsmtl.ca \\
	\addr ÉTS Montréal, Canada
 	\AND
        \firstname Ismail \surname Ben Ayed \email ismail.benayed@etsmtl.ca \\
	\addr ÉTS Montréal, Canada 
}

\begin{document}

\maketitle

\begin{abstract}
    Training a deep learning model to classify histopathological images is challenging, first, because of the color and shape variability of the cells and tissues, and second, because of the reduced amount of available data, which does not allow proper learning of those variations. Variations can come from the image acquisition process, for example, due to different cell staining protocols or tissue deformation.
    To tackle this challenge, Data Augmentation (DA) can be used during training to generate additional samples by applying transformations to existing ones. Those samples will help the model to become invariant to those color and shape transformations. The problem with DA is that it is not only dataset-specific but it also requires domain knowledge, which is not always available. Without this knowledge, selecting the right transformations can only be done using heuristics or through a computationally demanding search.
    To address this, we propose in this work an automatic DA learning method. In this method, the DA parameters, i.e. the transformation parameters needed to improve the model training, are considered learnable parameters and are learned automatically using a bilevel optimization approach in a quick and efficient way using truncated backpropagation.
    We validated the method on six different datasets of histopathological images. Experimental results show that our model can learn color and affine transformations that are more helpful to train an image classifier than predefined DA transformations. Predefined DA transformations are also more expensive as they need to be selected before the training by grid search on a validation set. We also show that similarly to a model trained with a RandAugment-based framework, our model has also only a few method-specific hyperparameters to tune but is performing better. This makes our model a good solution for learning the best data augmentation parameters, especially in the context of histopathological images, where defining potentially useful transformation heuristically is not trivial.
    Our code is available at~\url{https://github.com/smounsav/bilevel_augment_histo}.
\end{abstract}


\section{Introduction}
\begin{figure}[t]
      \centering
      \includegraphics[width=\linewidth,trim={0.6cm 0 0.6cm 0},clip]{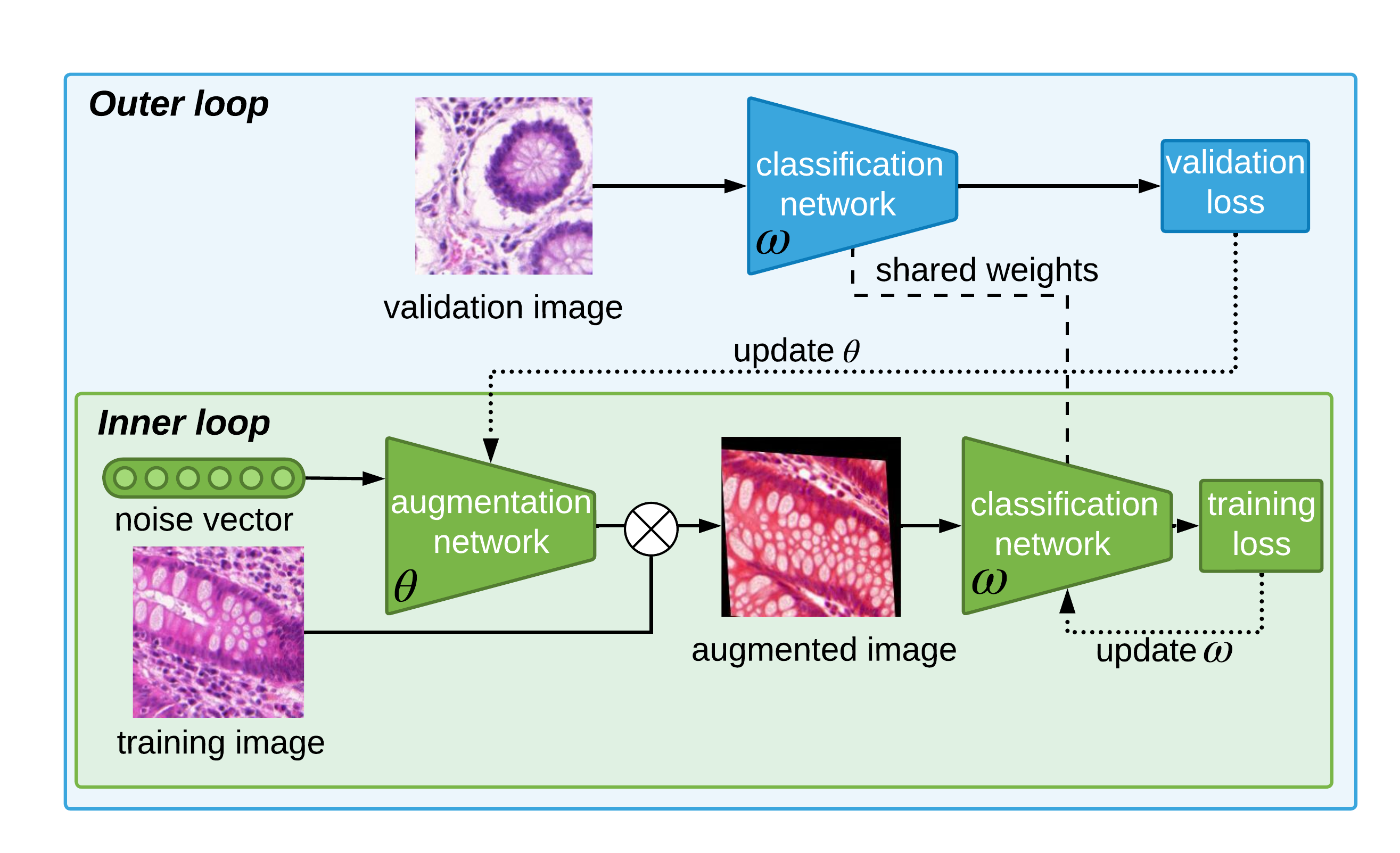}
      \caption{\textbf{Model training.} In an epoch, the classifier parameters $\omega$ are trained on the training set in the inner loop in the standard supervised way. Then, in the outer loop, the parameters of the augmentation network generating the data augmentation parameters are trained on the validation set using an online differentiable method.}
      \label{fig:model}
      \vspace{-3mm}
    \end{figure}

    Deep learning-based models have proved effective for the analysis of histopathological images \citep{Bejnordi2016StainSS,Litjens2017,shen2017deep,ker2018deep}. In the context of image classification, one hurdle to a good generalization is the color and shape variability of the cells and tissues in the images. Those variations can be inherent to the image acquisition process. More precisely, color variations can come from the cell staining, which is done to make the cells visible to the human eyes, whereas shape variations can come from tissue deformation. \\
    As those variations have a major impact on the performance of the models, addressing this issue has been an active area of research~\citep{Ciompi2017TheIO,Tellez2019QuantifyingTE,ataky2020data,faryna2021tailoring,electronics10050562,wagner2021structure,GARCEA2022106391}.
    To address color variations problems, two main directions can be followed: stain normalization and data augmentation. Stain normalization consists in altering the color space of the input images so that the difference between the color statistics of the train images and the color statistics of the test images is reduced. Data augmentation consists in creating new images from existing samples with different transformations so that the model can learn transformation invariances. In the context of histopathological image classification, data augmentation is particularly interesting as it can address the problem of both color and shape variations by teaching the classification model to be invariant to those two types of transformations.
    While data augmentation has been explored extensively for natural images as shown in \cite{Shao2022ATO}, most methods proposed rely on the selection of data augmentation parameters based on heuristics. Selecting meaningful transformations and the right amplitude requires prior or expert knowledge, which is not always available, especially in the medical images field \citep{Tellez2019QuantifyingTE}. Without the adequate knowledge, selecting the right transformations is not trivial, and selecting the wrong transformations can lead to a degradation of the model performance as shown in \cite{chen2020group}.
    To tackle the problem of selecting the right transformations, automatic data augmentation methods based on bilevel optimization like \cite{Cubuk_2019_CVPR} have been proposed for natural images. Those methods can be computationally expensive as for each new set of parameters, the model in the inner loop needs to be fully trained till convergence. Those methods were improved using a gradient-based approach like in \cite{Hataya2019FasterAL, Lin2019OnlineHL}, or more recently \cite{Hataya_2022_WACV}. In our work, as we can see in~Fig.\ref{fig:model}, we do not need to train the classifier in the inner loop until convergence for each different set of data augmentation parameters to test. The augmenter network generating the right data augmentation is trained at the same time as the classifier by alternating between the outer and the inner loop at each iteration.
    
    To address the problem of selecting the right data augmentation transformations, we propose in this work to extend the method in \cite{Mounsaveng2021LearningDA} to histopathological images. The method consists in training an image classifier while learning the best data augmentation transformations in a bi-level optimization framework. The classifier is trained in the inner loop while the best data augmentation parameters are learned in the outer loop. To make the method computationally efficient, the gradient of the validation loss used to update the data augmentation parameters is estimated using truncated backpropagation and with only one iteration of the inner loop.
    We validated this method on six different histopathological images datasets. Experimental results show that our method yields a better final accuracy than predefined data augmentation found by grid search on a validation set. During training, it finds transformations in the color and affine transformation space that help the most the learning and is also less expensive to train than predefined data augmentation. As we can see in~Fig.\ref{fig:model}, the best augmentation parameters are learned at the same time as the classifier is trained by alternating between the outer and inner loop at each iteration whereas when we do a grid search, the classifier needs to be trained till convergence in the inner loop for each set of different data augmentation parameters to test. What is also interesting to note is that the learned transformations do not hurt the model performance when useful transformations are more challenging to find.
    Moreover, similarly to a model trained with RandAugment, our model requires only a few model-specific hyperparameters to tune (in our case the hyperparameters of the augmenter network) but shows a better final classification accuracy. Our intuition is that even if RandAugment based methods have proven efficient, learning the best transformations along the training can yield a better classification performance, as the timing when transformations are presented to the model is important as shown in \cite{Golatkar2019TimeMI}. 
    
    \paragraph{Contributions}
        We summarize our contributions as follows:\\
        \begin{itemize}
            \item We successfully extend an automatic data augmentation learning method to learn useful color and affine augmentations in the context of histopathological images. As this method is differentiable, we can efficiently optimize a large transformation network that learns to perform data augmentation automatically.
    
            \item We show that our proposed model learns different sets of transformations and achieves comparable or better results than hand-defined transformations or RandAugment based methods on six different datasets.  We also show that our model never learns transformations that hurt the model performance and avoids the problem of badly hand-chosen transformations.
        \end{itemize}
\section{Related Works}
    Generative Adversarial Networks (GANs)~\citep{Goodfellow2014GenerativeAN} can generate realistic new samples of a certain dataset or class, thus they can be adapted for data augmentation.
    \citet{mirza2014conditional} and \citet{DBLP:conf/icml/OdenaOS17} proposed to generate images conditioned on their class that could be directly used to augment a dataset.  Also based on GAN, but directly used for data augmentation, DAGAN~\citep{antoniou2018augmenting} conditions the augmented image on the input image.
    TripleGAN~\citep{chongxuan2017triple} and Bayesian data augmentation~\citep{tran2017bayesian} train a classifier jointly with the generator.
    These approaches generate general image transformations, but in practice, it is not as performant as using predefined transformations.
    TANDA~\citep{ratner2017learning} is the only GAN-based approach that uses predefined transformations. It defines a large set of transformations and learns how to combine them to generate new samples that follow the same distribution as the original data. In medical images, \cite{FridAdar2018SyntheticDA} use a GAN to generate new CT-Scan images to improve the training of a liver lesion classifier.
    
    Our model is more efficient than GAN based models, as it does not require to learn a separate model before training the classifier. It learns the best transformation parameters and classifier at the same time.

    \subsection{AutoAugment}
        AutoAugment~\citep{Cubuk_2019_CVPR} is a data augmentation method that learns sequences of transformations that maximize the classifier accuracy on a validation set. This objective is better than simply reproducing the same data distribution as in GAN-based models, as it favors transformations that generalize well on unseen data. 
        However, it is computationally expensive as it performs the complete bilevel optimization by training the classifier in the inner loop until convergence for each set of evaluated transformations. 
        Some solutions to reduce the computational cost were proposed in follow-up works. Fast AutoAugment~\citep{lim2019fast} optimizes the search space by matching the density between the training set and the augmented data. Alternatively, Population Based Augmentation (PBA)~\citep{ho2019pba} focuses on learning the optimal augmentation schedule rather than only the transformations. However, even if these approaches reduce the computational cost of AutoAugment, they do not leverage gradient information. Faster AutoAugment~\citep{Hataya2019FasterAL} does this by combining AutoAugment with a GAN discriminator and considering transformations as differentiable functions. OHL-Auto-Aug \citep{Lin2019OnlineHL} uses an online bilevel optimization approach and the REINFORCE algorithm on an ensemble of classifiers to estimate the gradient of the validation loss and learn an augmentation probability distribution.
        RandAugment~\citep{Cubuk2019RandaugmentPA} goes further by showing that the same performance level as AutoAugment can be obtained by randomly selecting transformations from a predefined pool and just tuning the number of transformations to use and a global (same for all transformations) magnitude factor. However, this approach also requires prior knowledge of useful transformations.
        In histopathological images, \cite{faryna2021tailoring} use the RandAugment method with a set of transformations extended with some specific color transformations. This leads to an improved performance of the classifier.
        
        Our model is more efficient than search-based methods as the data augmentation parameters are updated at each training iteration using the gradient of the validation loss obtained in the inner loop. This gradient is estimated using truncated backpropagation, which removes the need to train the model until convergence for each set of evaluated transformations.
    
    \subsection{Hyperparameter Learning}
        Our work has some roots in the hyperparameter optimization field, as data augmentation parameters can be considered as hyperparameters to tune.
        Hyperparameter tuning is essential to obtain the optimal performances when training neural networks on a given dataset. Classic approaches assume that the learning model is a black-box and use methods like grid search, random search~\citep{NIPS2011_4443, bergstra2013making}, Bayesian optimization~\citep{snoek12practical}, or a tree-search approach~\citep{hutter2011sequential}. These approaches are simple but expensive because they repeat the optimization from scratch for each sampled value of the hyperparameters and so are only applicable to low dimensional hyperparameter spaces. 
        A different line of research is to leverage the gradient of these (continuous) hyperparameters (or hyper-gradients) to perform the hyper-optimization.
        The first work proposing this idea~\citep{bengio2000gradient} shows that the implicit function theorem can be used to this aim. This idea was developed more recently in \cite{bertrand2020implicit}.
        \cite{domke2012generic} was the first work to propose a gradient-based method using the bilevel optimization approach proposed in \cite{colson2007overview} to learn hyperparameters. Using a bilevel optimization approach to train a neural network is challenging, as usually there is no closed-form expression of the function learned in the inner loop (Section~\ref{sec:learning}).
        To address this, \citet{pmlr-v37-maclaurin15} and later \citet{pmlr-v70-franceschi17a} proposed methods to reverse the forward pass to compute the gradient of the validation loss. However, these methods are applicable only when the number of hyperparameters and the complexity of the models are limited due to the memory needed to save the intermediate steps.
        Another approach to address the computational hurdle in the inner loop is to calculate an approximation of the gradient like in \citet{pmlr-v48-pedregosa16} \citet{luketina2016scalable} or \citet{mackay2019self}. Our method differentiates from those by using truncated backpropagation to estimate the gradient of the validation loss.
        Finally, note that hyperparameter optimization presents some similarities to meta-learning as shown in \citet{pmlr-v80-franceschi18a}. For instance, in MAML~\citep{finn2017model}, a shared model initialization is learned to minimize the validation loss and therefore improve the generalization capabilities of the model. More recently, \cite{Hataya_2022_WACV} can be positioned at the intersection of AutoAugment and meta-learning based approaches.
            
        \begin{figure*}[t]
            \captionsetup[subfigure]{aboveskip=5pt,belowskip=-3pt}
            \centering
            \begin{subfigure}[t]{0.5\textwidth}
                \includegraphics[height=7cm,trim={0 4cm 0 4cm}]{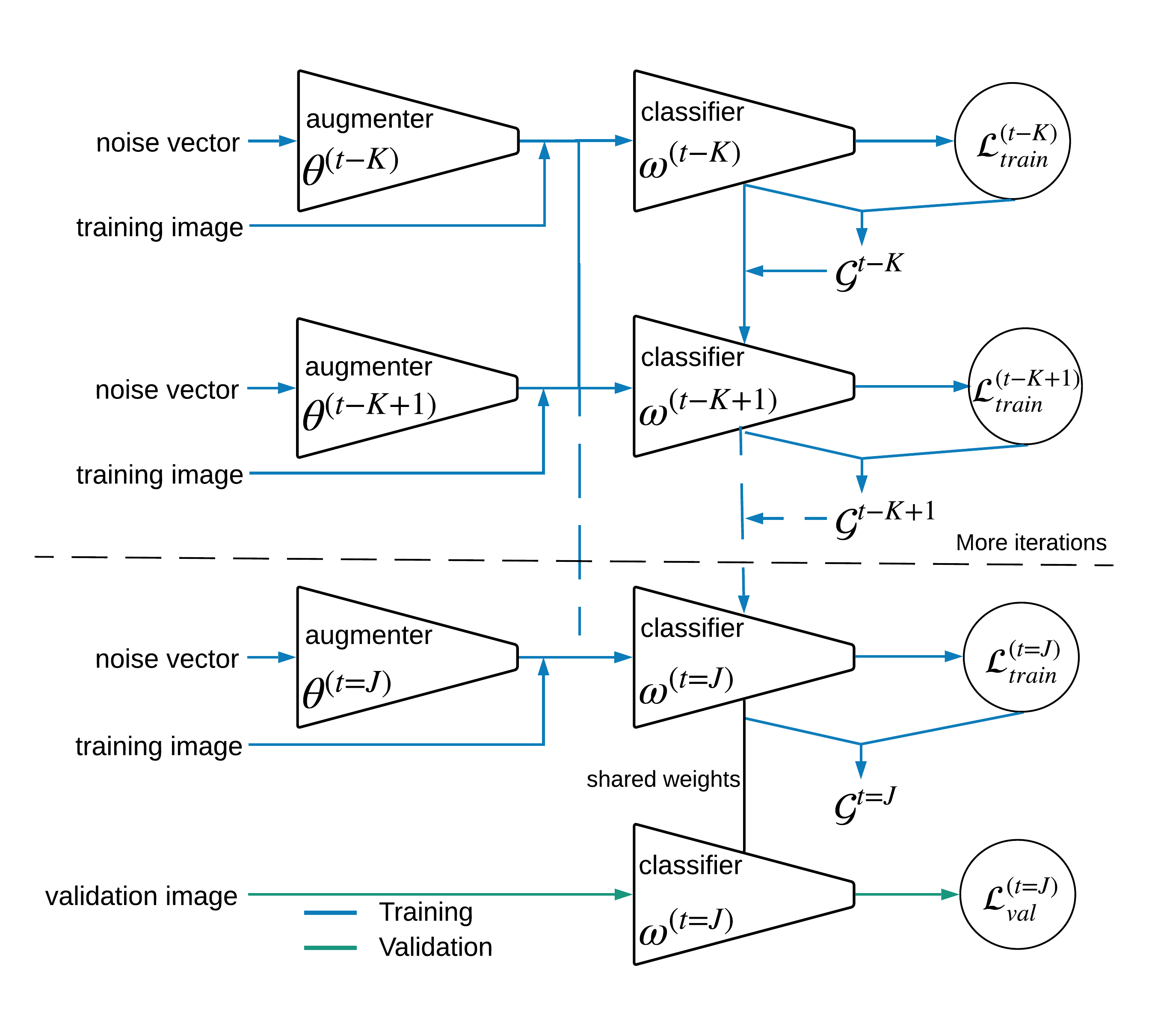}
                \caption{Forward pass.}\label{fig:forward}
            \end{subfigure}
                \hspace*{\fill}
                \begin{subfigure}[t]{0.4\textwidth}
                \includegraphics[height=7cm,trim={0 6cm 0 5cm}]{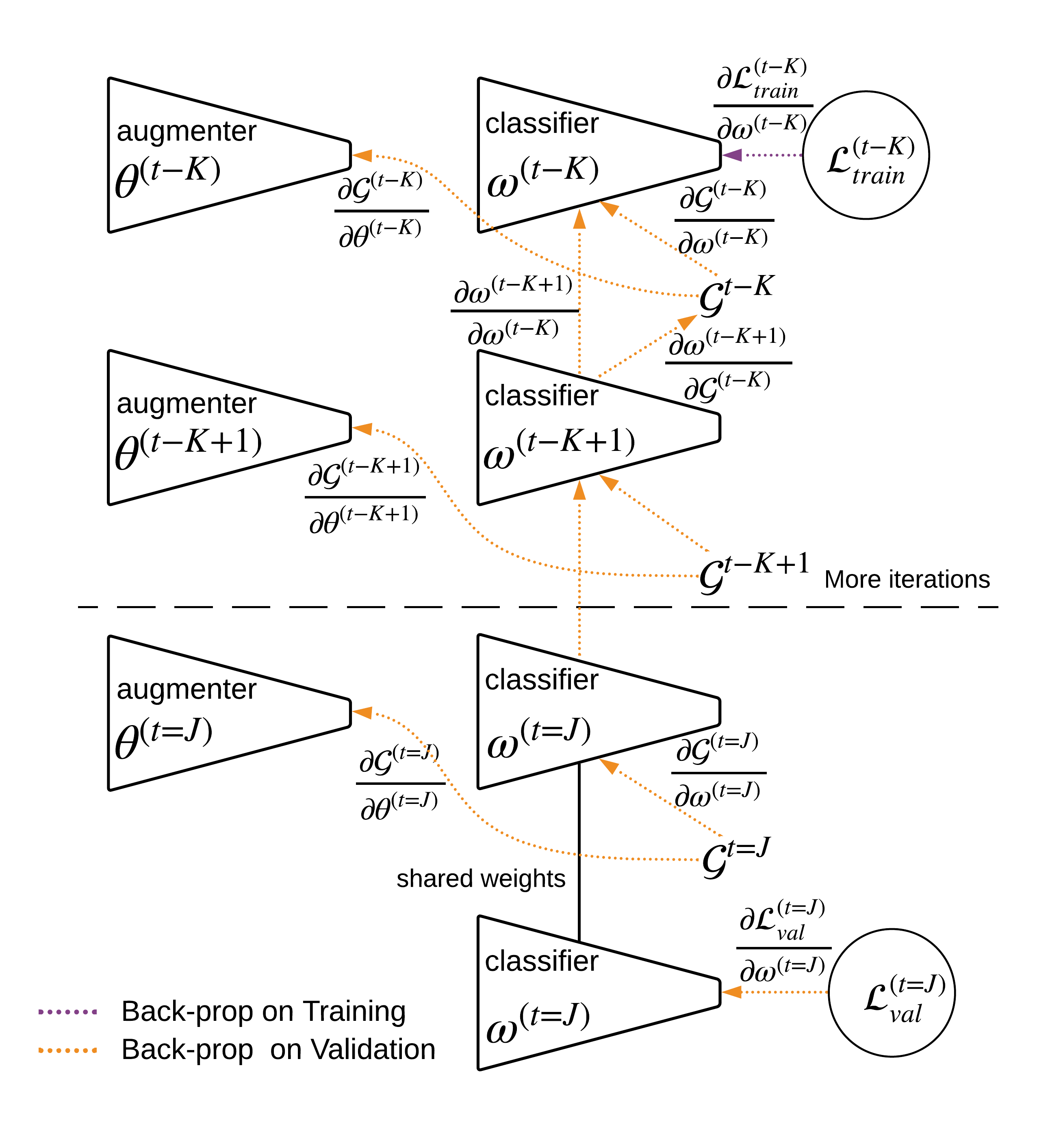}
                \caption{Backward pass.}\label{fig:backward}
            \end{subfigure}
            \caption{\textbf{Computational graph of our model at iteration $\boldsymbol{t=J}$.}
            $K$ is the number of gradient unfolding steps, and J is the number of inner loop iterations after which $\theta$ gets updated. The case where K=J=T (T being the iteration of the classifier convergence) is the complete bilevel optimization as in Eq.\ref{equ:bilevel1} whereas K=J=1 corresponds to updating $\theta$ at each mini-batch ($K=1)$, using only one step of gradient unfolding ($J=1$).}
            \label{fig:graph}
        \end{figure*}

\section{Proposed Data Augmentation Method}
    \label{sec:learning}
    Consider a labeled set $\mathcal{X}:=\{x_i,y_i\}^{N}_{i=1}$, where $x_i$ is an input image, $y_i$ the associated class label, N the number of samples and $\hat{\mathcal{X}}$ the set of transformed images. We formulate the problem of identifying effective data augmentation transformations as a bilevel optimization problem. In this setup, the augmenter $\mathcal{A}_\theta:\mathcal{X}\rightarrow{\hat{\mathcal{X}}}$ is parametrized by ${\theta}$ and is used to minimize the loss $\mathcal{L}$ on the validation data $\mathcal{X}_{val}$ in the outer loop. In the inner loop, the classifier parameters $\omega$ are optimized  on the training data $\mathcal{X}_{tr}$ in the standard supervised way. This formulation can be written as:
    \begin{align}
        \theta^* &= \argmin_\theta
        \mathcal{L}(\mathcal{X}_{val}, \omega^*) 
        \label{equ:bilevel1} \\
        s.t. \quad \omega^* &= \argmin_{\omega} \mathcal{L}(\mathcal{A}_{\theta}(\mathcal{X}_{tr}),\omega).
        \label{equ:bilevel2}
    \end{align}
    While optimizing a few hyperparameters on the validation data is feasible with black-box approaches such as grid and random search~\cite{bergstra12random} or Bayesian optimization~\cite{snoek12practical}, it is not efficient. With bilevel optimization, our aim is to efficiently learn an entire neural network $\mathcal{A}_\theta$ (possibly with thousands of parameters $\theta$) which defines a distribution of transformations that should be applied on the training data to improve generalization. 
    
    Gradient descent was shown to be an efficient method for optimizing parameters of large networks. In problems such as architecture search~\citep{liu2018darts}, the parameters can be directly optimized with gradient descent (or second order methods) against the training and validation data.
    However, this is not the case for data augmentation. The reason is that the transformation network $\mathcal{A}_{\theta}$ is optimized to maximize the validation score, but applies transformations only on the training set. Therefore, first order methods would not work. The aim of data augmentation is to introduce transformations during the training phase that can make the model invariant or partially invariant to any transformations that can occur at test time. If we optimize the transformation network directly on the validation data, the model will simply select trivial solutions such as the identity transformation. This approach has been used for  object localization~\cite{jaderberg2015spatial} and it did not improve the model generalization performance as much as data augmentation.
    To solve this issue, new methods relied on reinforcement learning instead of gradient descent to learn effective data augmentation~\citep{Cubuk_2019_CVPR, lim2019fast, ho2019pba}.
    
    In this work, we show that in the case of a differentiable augmenter $\mathcal{A}_\theta$, there is a simple, efficient way to find optimal data transformations based on gradient descent that generalize well on validation data. We formulate our problem as an approximation to bilevel optimization by using truncated backpropagation as it allows our method to:
    i) efficiently estimate a large number of parameters to generate the optimal data augmentation transformations by gradient descent;
    ii) obtain an online estimation of the optimal data augmentation during the different phases of the training, which can also be beneficial~\citep{Golatkar2019TimeMI};
    iii) change the training data to adapt to different validation conditions as in supervised domain adaptation.\\
    Although approximate bilevel optimization has already been proposed for hyperparameter optimization~\citep{Shaban-AISTATS-19, pmlr-v80-franceschi18a, pmlr-v70-franceschi17a}, in this paper we show that it can be used for training a large, complex model (the augmenter $\mathcal{A}_\theta$ network) to learn an effective distribution of transformations.

    \subsection{Approximate Online Bilevel Optimization}
        As shown in Eq.~\ref{equ:bilevel1} and \ref{equ:bilevel2}, the problem of finding the optimal data augmentation transformations $\mathcal{A}_\theta$ can be cast as a bilevel optimization problem. This problem can be solved by iteratively solving Eq.~\ref{equ:bilevel2} to find the optimal network weight $\omega^*$, given the parameters of the transformation $\theta$ and then updating $\theta$:
        \begin{equation}
            \footnotesize
            \theta \leftarrow \theta - \eta_\theta\nabla_\theta \mathcal{L}(\mathcal{X}_{val},\omega^*)
        \end{equation}
        where $\eta_\theta$ is the learning rate used to train the augmenter network.\\ 
        However, as the augmentations are to be applied only on the training dataset and not on the validation set, calculating $\frac{\partial{\mathcal{L}(\mathcal{X}_{val},\omega^*)} }{\partial \theta}$ is not trivial. To enable this calculation, we use the fact that the weights $\omega$ of the network are shared between training and validation data and use the chain rule to differentiate the validation loss $\mathcal{L}(\mathcal{X}_{val},\omega^*)$ with respect to the hyperparameters $\theta$. In other words, instead of using a very slow black-box optimization for $\theta$, we can exploit gradient information 
        because the model parameters $\omega^*$ are shared between the validation and the training loss.\\
        We define the gradient of the validation loss with respect to $\theta$ as follows:
        \begin{equation} \label{equ:grad1}
            \footnotesize
            \begin{split}
                {\nabla_\theta \mathcal{L}}(\mathcal{X}_{val},\omega^*)&=\frac{\partial{\mathcal{L}(\mathcal{X}_{val},\omega^*)} }{\partial \theta}\\
                &= \frac{\partial{\mathcal{L}}(\mathcal{X}_{val},\omega^*)}{\partial \omega^*} \frac{\partial{\omega^*}}{\partial \theta}
            \end{split}
        \end{equation}
        By defining $\mathcal{G}^{(t)}$ as the gradient of the training loss at iteration $t$:
        \begin{equation}
            \footnotesize
            \mathcal{G}^{(t)}=\nabla_{\omega}\mathcal{L}(\mathcal{A}_{\theta}(\mathcal{X}_{tr}),\omega^t)
        \end{equation}
        we can write $\frac{\partial{\omega^*}}{{\partial \theta}}$ in Eq.~\ref{equ:grad1} as:
        \begin{equation}
            \footnotesize
            \frac{\partial{\omega^*}}{\partial \theta} = \sum_{i=1}^{T-1} \frac{\partial{\omega^{(T)}}}{\partial \omega^{(i)}} \frac{\partial{\omega^{(i)}}}{\partial{\mathcal{G}^{(i-1)}}}\frac{\partial{\mathcal{G}^{(i-1)}}}{\partial {\theta}}
        \end{equation}
        where T is the iteration when the classifier converges.
        
        As $\omega^*$ represents the model weights at training convergence, they depend on $\theta$ for each iteration of gradient descent. Thus, to compute $\frac{\partial{\omega^*}}{\partial \theta}$, one has to back-propagate throughout the entire $T$ iterations of the training cycle. An example of this approach is in \citet{pmlr-v37-maclaurin15}.
        This approach is feasible only for small problems due to the large requirements in terms of computation and memory. 
        However, as optimizing $\omega^*$ is an iterative process, instead of computing $\frac{\partial{\omega}}{\partial \theta}$ only at the end of the training loop, we can estimate it at every iteration $t$: 
        \begin{equation}
            \footnotesize
            \frac{\partial{\omega^*} } {\partial \theta} \approx
            \frac{\partial{\omega^{(t)} } } {\partial \theta^{(t)}} = \sum_{i=1}^{t} \frac{\partial{\omega^{(t)}}}{\partial{\omega^{(i)}}} \frac{\partial{\omega^{(i)} }} {\partial{\mathcal{G}^{(i-1)}}}\frac{\partial{\mathcal{G}^{(i-1)}}}{\partial {\theta^{(i)}}},
            \label{equ:online}
        \end{equation}
        This procedure corresponds to dynamically changing $\theta$ during the training iterations (thus it becomes $\theta^{(t)}$) to minimize the current validation loss based on the training history. Although this formulation is different from the original objective function, adapting the data augmentation transformations dynamically with the evolution of the training process can improve generalization performance~\citep{Golatkar2019TimeMI}.
        This relaxation is often used in constrained optimization for deep models, in which constraints are reformulated as penalties and their gradients are updated online, without waiting for convergence, to save computation \citep{Pathak2015ConstrainedCN}. However, in our case, we cannot write the bilevel optimization as a single unconstrained formulation in which the constraint in $\omega^*$ is summed with a multiplicative factor that is maximized (i.e., Lagrange multipliers), because the upper level optimization should be performed only on $\theta$, while the lower level optimization should be performed only on $\omega$. Nonetheless, even with this relaxation, estimating $\frac{\partial{\omega^*} } {\partial \theta}$ still remains a challenge as it does not scale well. Indeed, the computational cost of computing $\frac{\partial{\omega^{(t)}}}{\partial{\theta^{(t)}}}$ grows with the number of iterations $t$ as shown in Eq.~\ref{equ:online}.
        To make the gradient computation constant at each iteration we use truncated backpropagation similarly to what is commonly used in recurrent neural networks~\citep{Williams90anefficient}:
        \begin{equation} \label{equ:truncated}
            \footnotesize
            \frac{\partial{\omega^{(t)} } } {\partial \hat\theta} \approx \sum_{i=t-K}^{t} \frac{\partial{\omega^{(t)} } } {\partial{\omega^{(i)} } } \frac{\partial{\omega^{(i)} }} {\partial{\mathcal{G}^{(i-1)}}}\frac{\partial{\mathcal{G}^{(i-1)}}}{\partial {\theta^{(i)}}},
        \end{equation}
        where $K$ represents the number of gradient unfolding that we use. 
        Fig.~\ref{fig:graph}b. shows the computational graph used for this computation.\\
        Additionally, as \citet{Williams90anefficient}, we consider a second parameter $J$ which defines the number of inner loop training iterations after which $\theta$ is updated, in other words how often the computation of the gradients of $\theta$ is performed. The situation where $K=J=T$ is the exact bilevel optimization as shown in Eq.~\ref{equ:bilevel1} while $K=J=1$ corresponds to updating $\theta$ at each iteration, in our case mini-batch ($K=1$), using only one step of gradient unfolding ($J=1$). A theoretical analysis of the convergence of this approach is presented in \citet{Shaban-AISTATS-19}.

    \subsection{Augmenter Network}
        In this work, we use an augmenter network that can learn two types of transformations: geometrical and color. We use the transformation model of spatial transformer networks~\citep{jaderberg2015spatial}, but for data augmentation instead of data alignment. Thus, as illustrated in Fig.~\ref{fig:model}, the augmenter is composed of a module that generates a set of transformation parameters followed by a module that applies the generated transformations to the original image. Note that the learned transformations are not conditioned on the input image but defined only based on random noise.\\                
        \subsubsection{Geometrical} 
            In our experiments, we consider scenarios where the augmenter network learns affine transformations. The choice of this kind of transformation is motivated by the fact that tissues are deformed during the image acquisition process. By considering affine transformations in our learned data augmentation, we aim to train the model to become invariant to those geometrical deformations.
            For affine transformations, the augmenter network receives as input a random noise vector and generates a 2x3 matrix of values representing a variation from the identity transformation.

        \subsubsection{Color}
            We also consider scenarios where the augmenter learns color transformations. Color augmentations are important as they can help the trained model to become invariant to color perturbations appearing during the cell staining process.
            Color transformations considered are: contrast, brightness and in the HSV space hue and saturation. For color transformations, the augmenter receives as input a random noise vector and generates a single value representing a variation for each color transformation. In our implementation, we use the kornia library~\citep{Riba2019KorniaAO}, which follows the specifications of \cite{Szeliski2010ComputerV}.
            For contrast, the value learned is a non-negative factor applied to the actual color values. 1 represents the initial image whereas values tending to 0 mean a black-and-white image. If we consider the variables $r$, $g$, and $b$ representing the values of the red, green, and blue colors of the images and $cf$ the contrast factor learned by our network, the new RGB values are obtained using the update rule:
            \begin{equation}
                (r, g, b) \leftarrow clamp((r, g, b)\cdot cf, 0, 1)
            \end{equation}
            For brightness, the value learned represents a shift applied to the actual color values. 0 represents the initial image. If we consider the variables $r$, $g$, and $b$ representing the values of the red, green, and blue colors of the images and $bs$ the brightness shift learned by our network, the new RGB values are obtained using the update rule:
            \begin{equation}
                (r, g, b) \leftarrow clamp((r, g, b) + bs, 0, 1)
            \end{equation}
            In the case of saturation, the value learned by the augmenter is a non-negative factor applied to the actual saturation value. A value of 1 represents the original image whereas 0 means a black-and-white image. If we consider the variables $h$, $s$, and $v$ representing the values of the hue, saturation, and value of the images and $sf$ the saturation factor learned by our network, the new HSV values are obtained using the update rule:
            \begin{equation}
                (h, s, v) \leftarrow clamp((h, s\cdot sf, v) , 0, 1)
            \end{equation}
            Finally, the value learned by our augmenter for hue is a shift of the hue channel. 0 represents no shift to the hue channel and any other value negative or non-negative is added to the actual value. If we consider the variables $h$, $s$, and $v$ representing the values of the hue, saturation, and value of the images and $hs$ the hue shift learned by our network, the new HSV values are obtained using the update rule:
            \begin{equation}
                (h, s, v) \leftarrow (mod(h + hs, 2\pi), s, v)
            \end{equation}

    \section{Experimental setup}
        \subsection{Datasets and evaluation}
            \label{sec:datasets}
            The datasets used in our experiments are:

            \paragraph{BACH}
             \citet{Aresta2019BACHGC} is a dataset of 400 H\&E (hematoxylin and eosin) stained breast cancer histology images of resolution 2048 x 1536 distributed in 4 balanced classes of 100 images. As there is no test set publicly available, we use in our experiments 40\% of the dataset for training, 10\% for validation and 50\% for testing as in \cite{Rony2022DeepWL}. The values used for the predefined color transformations are brightness=0.5, contrast=0.5, saturation=0.5 and hue=0.05.

            \paragraph{Glas}
             \citet{Sirinukunwattana2016GlandSI} is a dataset of 165 H\&E stained colon cancer histology images of variable resolution (in our experiments, we use an image size of 430x430) distributed in 2 classes (benign and malignant). The dataset is divided in a train set of 85 images (37 benign and 48 malignant) and a test set of 80 images (37 benign and 43 malignant). In our experiments, we use 80\% of the training set for training and 20\% for validation. The values used for the predefined color transformations are brightness=0.25, contrast=0.25, saturation=0.25 and hue=0.4.

            \paragraph{HICL Larynx}
             \citet{Ninos2015MicroscopyIA} is a dataset of 450 H\&E and P63 stained larynx cancer histology images with 2 magnifying factors (20x and 40x). It has 3 classes corresponding to cancer grades: Grade I, II and III. For the 20x magnification factor, the image resolution is 1728x1296 and the number of images per class is I:87, II:73 and III:64. For the 40x magnification factor, the image resolution is 1300x1030 and the number of images per class is I:88, II:74 and III:64. As there is no test set publicly available, we use in our experiments 70\% of the dataset for training, 20\% for validation set, and 10\% for test. The values used for the predefined color transformations are brightness=0.25, contrast=0.25, saturation=0.25 and hue=0.4.
             
            \paragraph{HICL Brain}
             \citet{Glotsos2008ImprovingAI} is a dataset of 2548 H\&E and P63 stained brain cancer histology images with 2 magnifying factors (20x and 40x). It has 7 classes corresponding to cancer grades: Grade I, I-II, II, II-III, III, III-IV and IV. For the 20x magnification factor, the image resolution is 1728x1296 and the number of images per class is I:123, I-II: 94, II:208, II-III:47, III:367, III-IV:45 and IV:373. For the 40x magnification factor, the image resolution is also 1728x1296 and the number of images per class is I:132, I-II: 73, II:210, II-III:53, III:434, III-IV:32 and IV:357. As there is no test set publicly available, we use in our experiments 70\% of the dataset for training, 20\% for validation set, and 10\% for test. The values used for the predefined color transformations are brightness=0.25, contrast=0.25, saturation=0.25 and hue=0.4.

            \paragraph{Evaluation} To evaluate the performance of our models, we use the classification accuracy metric, which is defined by the number of samples correctly classified divided by the total number of samples. For each scenario, we do a 5-fold cross-validation and the result reported is the average of the results obtained by the 5 folds. The hyperparameters search is done separately for each dataset. The hyperparameters selected are the ones yielding the best validation results averaged over the 5 folds. We also follow this protocol for Randaugment hyperparameters.

        \subsection{Implementation details}
            \label{sec:implementation_details}
            As we can see in~Fig.\ref{fig:model}, our model is composed of a classifier and an augmenter network. As classifier, we use a ResNet18~\citep{He2015ResNet} network pretrained on Imagenet. ResNet18 is an 18 layers deep neural network with residual connections. To align with the image size used during pretraining, we use in our training phase patches of size 224x224 and evaluate the model on whole images during the testing phase.
            The augmenter learning the geometric and color transformations is a MLP network that receives a noise vector of dimension 100 as input and generates the transformation parameters.
            The augmenter network has 3 fully connected layers of size 100, 64, and 32 and an output layer of size $n$, $n$ being the number of hyperparameters to optimize according to the scenario considered (4 for color transformations, 6 for affine transformations, or 10 when both color and affine transformations are considered). 
            To have differentiable affine and color transformations, we use the affine\_grid and grid\_samples functions of the torchvision package of Pytorch framework~\citep{NEURIPS2019_9015}.
            
            As in \cite{Mounsaveng2021LearningDA}, we adopt a frequency K of updating $\theta$ $J$ of 1 and a number of steps of backpropagation J of 1 to update the parameters of our model. 

    \section{Results}
        Our proposed method aims at learning data augmentation automatically while training the image classifier. We validate our method on the 6 different datasets presented in \ref{sec:datasets}: BACH, Glas, Medisp HICL Larynx with magnification factor 20x, Medisp HICL Larynx with magnification factor 40x, Medisp HICL Brain with magnification factor 20x, Medisp HICL Brain with magnification factor 40x.
        
        In a first series of experiments, we compare for each dataset the best performance of an image classifier trained on 3 different kinds of transformations: first, we consider color transformations, then affine transformations, and finally a combination of both transformation types. For each type of transformation, we compare the classification performance of 3 models: first, a classifier trained without data augmentation (baseline), then a classifier trained with hyperparameters for data augmentation found by grid search on a validation set (predefined), and finally a classifier trained with the data augmentation learned by our method. Note that in all our experiments, we divide our data into 3 sets: a training set to train our model, a validation set to select the hyperparameters of the model, and a test set to evaluate the model.

        Then, in a second series of experiments, we investigate our model more in-depth focusing on the BACH dataset. We chose this dataset as it is the most challenging of the six considered in this paper in terms of image size and difficulty to learn the classification task. More precisely, we investigate two aspects of our model: i) the impact of the amount of training data available on the quality of the learned augmentations ii) the impact of starting the training from a classifier pretrained on Imagenet.

        Finally, in a third series of experiments, we compare our approach to a model trained in a data augmentation framework similar to RandAugment \citep{Cubuk2019RandaugmentPA}. Instead of searching for the best sequence of transformations and the best magnitude for each transformation at the same time as in other models of the AutoAugment family, RandAugment relaxes the search problem to the tuning of 2 hyperparameters M and N, M being a global magnitude for all considered transformations and N the number of transformations selected in each sequence of transformations. In our experiments, we define M and N by doing a grid search with values between 1 and 5 for both M and N. This approach is simple yet very efficient and has proven to be state of the art in \cite{faryna2021tailoring} on Camelyon 17 dataset. However, we argue that even if RandAugment is very simple and efficient to use, it still requires prior knowledge to define the initial pool of transformations. For our method, we also need to fine-tune a limited number of hyperparameters (the hyperparameters of the augmenter network), but we can also define a more generic set of differentiable transformations. Moreover, learning the optimal data augmentation at each epoch can be beneficial for the model, as the time when the transformations are presented to the model is important as reported in \cite{Golatkar2019TimeMI}.
        To be fair in the comparison of the results, we limited the pool of transformations used by RandAugment to the transformations learned by our proposed model. The transformations considered by our adapted RandAugment framework are listed in Tab. \ref{tab:randaugment_transformation_set}.

        \begin{table}[t]
            \footnotesize
            \centering
            \begin{tabular}{c|c}
                Transformation type & Magnitude Range\\
                \hline
                identity & - \\
                rotation & [-30.0, 30.0]\\
                translation x & [-0.45, 0.45]\\
                translation y & [-0.45, 0.45]\\
                shear x & [-0.3, 0.3]\\
                shear y & [-0.3, 0.3]\\
                contrast & [0, 2]\\
                brightness & [-1, 1]\\
                hue & [-0.5, 0.5]\\
                Saturation & [0, 1]\\
            \end{tabular}%
            \caption{\textbf{Transformations considered by our adapted RandAugment framework}. To be fair in the comparison with our proposed model, we limited the set of transformations to the differentiable transformations learned by our model.}
            \label{tab:randaugment_transformation_set}
        \end{table}

        \begin{table*}[t]
            \small
            \centering
            \setlength\tabcolsep{2pt}
            \begin{tabular}{l|c|c|c|c|c|c}
                Scenario / Dataset &  BACH & Glas & \makecell{Larynx\\20x} & \makecell{Larynx\\40x} & \makecell{Brain\\20x} & \makecell{Brain\\40x}\\
                \hline
                Baseline & 83.30\tiny$\pm$1.18 & 89.50\tiny$\pm$1.22 & 87.51\tiny$\pm$1.27 & 86.67\tiny$\pm$1.16 & 99.53\tiny$\pm$0.43 & 96.97\tiny$\pm$1.13\\
                Baseline + color DA & 85.10\tiny$\pm$1.19 & 96.00\tiny$\pm$1.25 & 90.24\tiny$\pm$1.38 & 86.67\tiny$\pm$1.23 & 99.53\tiny$\pm$0.43 & 97.42\tiny$\pm$1.20\\
                Baseline + affine DA & 83.70\tiny$\pm$1.20 & 97.75\tiny$\pm$1.27 & 95.92\tiny$\pm$1.11 & 94.29\tiny$\pm$1.38 & 99.54\tiny$\pm$0.68 & 98.18\tiny$\pm$1.05\\                        
                Baseline + color\&affine DA & 84.60\tiny$\pm$1.23 & 98.25\tiny$\pm$1.12 & 95.24\tiny$\pm$1.25 & 95.24\tiny$\pm$1.37 & 99.53\tiny$\pm$0.43 & 98.94\tiny$\pm$1.26\\    
                \hline    
                Our model (color DA) & 85.60\tiny$\pm$1.18 & 97.25\tiny$\pm$1.23 & 91.11\tiny$\pm$1.28 & 86.67\tiny$\pm$1.22 & 99.53\tiny$\pm$0.43 & 97.57\tiny$\pm$1.24\\
                Our model (affine DA) & 85.40\tiny$\pm$1.25 & 98.25\tiny$\pm$1.20 & 96.19\tiny$\pm$1.22 & 95.24\tiny$\pm$1.26 & 99.69\tiny$\pm$0.43 & 98.63\tiny$\pm$1.36\\
                Our model (color\&affine DA) & \textbf{88.90}\tiny$\pm$1.25 & \textbf{99.25}\tiny$\pm$0.56 & \textbf{96.61}\tiny$\pm$1.23 & \textbf{97.14}\tiny$\pm$1.26 & \textbf{99.84}\tiny$\pm$0.35 & \textbf{99.24}\tiny$\pm$0.54\\
            \end{tabular}%
            \caption{\textbf{Impact of color and affine transformations on classification accuracy (\%)}. Transformations in parentheses are learned, others are predefined. Our model performs better than hand-defined transformations on the six different datasets. Best performances are obtained with a combination of learned color and affine transformations.}
            \label{tab:experiments_results}
        \end{table*}        

        \subsection{Color Transformations}
            In this section, we investigate the impact of color transformations alone on the training of an image classifier.
            Results are presented in Tab.\ref{tab:experiments_results}.
            For BACH dataset, using predefined color augmentations to train the model yields an increased classification performance compared to the baseline, but the model has the best performance when trained with the augmentations learned by our augmenter (+2.3\% accuracy VS baseline and +0.5\% accuracy VS predefined color augmentations). 
            For Glas dataset, the classifier performs better with learned transformations than with predefined ones (+7.75\% over baseline and +1.25\% over baseline). 
            For Larynx 20x dataset, our model also performs better than predefined augmentations (+3.6\% accuracy VS baseline and +0.87\% accuracy VS predefined augmentations). 
            For Larynx 40x dataset, our model performs similarly to predefined transformations. However, in both cases, we do not see any improvement over the baseline, which indicates that either color transformations might not be the best ones to use for this dataset or that the performance of the classifier is already saturated to see the improvement brought by those transformations. 
            For Brain 20x dataset, similarly to the Larynx 20x dataset, our model performs on-par with predefined transformations and brings no improvement over the baseline. Also in this case, it seems that color augmentations used are not useful to train the classifier or that the performance is too saturated to see the improvement.
            For Brain 40x dataset, our model performs slightly better than predefined augmentations (+0.6\% accuracy VS baseline and +0.13\% accuracy over predefined data augmentations. 

        \subsection{Geometric Transformations}
            In this section, we evaluate our model by investigating the impact of geometric transformations on the classification accuracy.
            Results are presented in Tab.\ref{tab:experiments_results}.
            For BACH dataset, our model performs better than predefined affine transformations (+2.1\% accuracy over baseline and +1.7\% over predefined transformations). However, it does not perform as well as when learning only color transformations, which indicates that this kind of transformations is less efficient for this dataset.
            For Glas dataset, our model performs also better than predefined affine transformations (+8.75\% accuracy over baseline and +0.5\% over predefined transformations). 
            Interesting to note is also that for this dataset, affine transformations are helping more to train the model than using only color transformations.
            For Larynx 20x dataset, we can see that our model performs slightly better than predefined transformations (+8.68\% accuracy over baseline and +0.27\%  over predefined transformations). 
            Also for this dataset, using geometric transformations seems to help train the model more than using color transformations only.
            For Larynx 40x dataset, similarly to Larynx 20x dataset, our model performs slightly better (+8.57\% accuracy over baseline and +0.95\% over predefined augmentations) and affine transformations are more helpful than only color transformations. 
            For Brain 20x dataset, our model performs slightly better than predefined affine transformations (+0.16\% accuracy over baseline and +0.15\% over predefined affine transformations. As opposed to color transformations, affine transformations have a positive impact on the performance of the classification model. 
            For Brain 40x dataset, our model performs also slightly better than predefined affine transformations (+1.66\% accuracy over baseline and +0.45\% over predefined augmentations). 
            Also for this dataset, affine transformations have a bigger positive impact on the model accuracy than color transformations.
            
        \subsection{Combination of color and affine transformations}
            In this section, we evaluate our model by investigating the impact of geometric transformations on the classification accuracy.
            Results are presented in Tab.\ref{tab:experiments_results}.
            For BACH dataset, the combination of both kind of transformations is significantly improving the classification score (+5.6\% accuracy over baseline and +4.3\% over predefined augmentations). 
            For Glas dataset also, the combination of both color and geometric transformations is improving the model performance (+9.75\% accuracy over baseline and +1\% over predefined augmentations). 
            For Larynx 20x dataset, the combination of color and affine transformations learned by our model has a bigger positive impact on the model final accuracy (+9.1\% over baseline and +1.37\% over predefined transformations). 
            For Larynx 40x dataset, learning both color and affine transformations is also yielding the model with the best classification accuracy (+10.47\% over baseline and +1.9\% over predefined transformations). 
            For Brain 20x dataset, our model learning color and affine transformations is performing slightly better than predefined transformations (+0.31\% over baseline and predefined transformations). 
            For Brain 40x dataset, when learning color and affine transformations at the same time our model is performing better than the predefined transformations (+2.27\% over baseline and +0.3\% over predefined transformations). 
            To summarize the results of this series of experiments, we can see that the combination of both color and affine transformations yields the best results, which shows that our model became more invariant to color and shape perturbations thanks to the data augmentation transformation learned along the training.

        \subsection{Additional experiments on BACH dataset}

            In this section, we run a series of experiments on BACH dataset to have a better understanding of our model. BACH was chosen as it is the most challenging dataset of the six considered in terms of image size and difficulty of the classification task as shown in Tab.~\ref{tab:experiments_results}.

            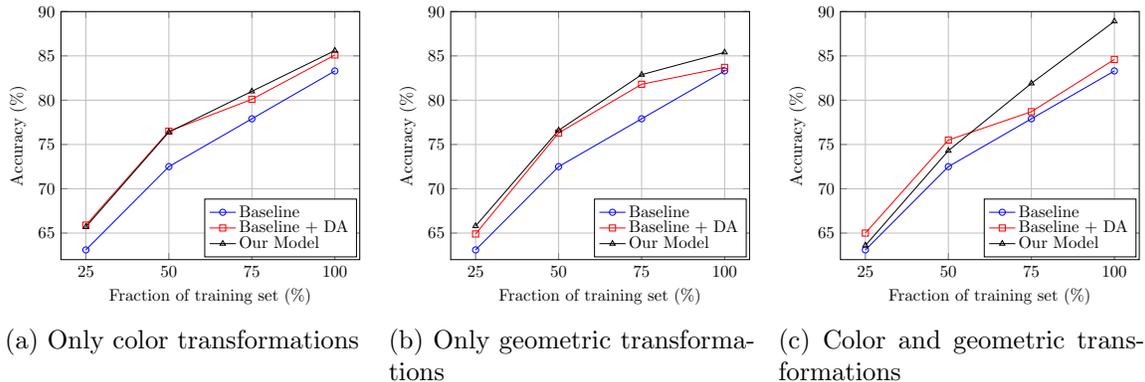
\begin{figure}[t]
                \centering
                \begin{subfigure}[t]{0.32\columnwidth}
    \centering
    \resizebox{\columnwidth}{!}{
        \begin{tikzpicture}[scale=1.0]
          \begin{axis}[ 
            legend cell align={left},
            legend style={legend style={row sep=-3pt},
            at={(0.99,0.01)},anchor=south east},
            xlabel=Fraction of training set (\%),
            ylabel=Accuracy (\%),
            grid=major,
            xlabel near ticks,
            xticklabel style={/pgf/number format/1000 sep=},
            ylabel near ticks,
            xtick=data,
            yticklabel style={
                left,
                /pgf/number format/.cd,
                fixed,
                precision=2,
                /tikz/.cd
            },
            log ticks with fixed point,
            ymin=62,
            ymax=90,
          ]
            \addplot[color=blue,mark=o] coordinates { (25, 63.1) (50,72.5) (75,77.9) (100,83.3)};
            \addplot[color=red,mark=square] coordinates { (25,65.9) (50,76.5) (75,80.1) (100,85.1)};
            \addplot[color=black,mark=triangle] coordinates { (25,65.7) (50,76.4) (75,81) (100,85.6)};                                
            \legend{Baseline, Baseline + DA, Our Model}
          \end{axis}
        \end{tikzpicture}
    }
    \caption{Only color transformations}
    \label{fig_BACH_color}
\end{subfigure}
                \hfill    
                \begin{subfigure}[t]{0.32\columnwidth}
    \centering
    \resizebox{\columnwidth}{!}{
        \begin{tikzpicture}[scale=1.0]
          \begin{axis}[ 
            legend cell align={left},
            legend style={legend style={row sep=-3pt},
            at={(0.99,0.01)},anchor=south east},
            xlabel=Fraction of training set (\%),
            ylabel=Accuracy (\%),
            grid=major,
            xlabel near ticks,
            xticklabel style={/pgf/number format/1000 sep=},
            ylabel near ticks,
            xtick=data,
            yticklabel style={
                left,
                /pgf/number format/.cd,
                fixed,
                precision=2,
                /tikz/.cd
            },
            log ticks with fixed point,
            ymin=62,
            ymax=90,
            cycle list name=color
          ] 
            \addplot[color=blue,mark=o] coordinates { (25, 63.1) (50,72.5) (75,77.9) (100,83.3)};
            \addplot[color=red,mark=square] coordinates { (25,64.9) (50,76.3) (75,81.8) (100,83.7)};
            \addplot[color=black,mark=triangle] coordinates { (25,65.8) (50,76.6) (75,82.88) (100,85.4)};                    
            \legend{Baseline, Baseline + DA, Our Model}
          \end{axis}
        \end{tikzpicture}
    }
    \caption{Only geometric transformations}
    \label{fig_BACH_geo}
\end{subfigure}
                \hfill
                \begin{subfigure}[t]{0.32\columnwidth}
    \centering
    \resizebox{\columnwidth}{!}{
        \begin{tikzpicture}[scale=1.0]
          \begin{axis}[ 
            legend cell align={left},
            legend style={legend style={row sep=-3pt},
            at={(0.99,0.01)},anchor=south east},
            xlabel=Fraction of training set (\%),
            ylabel=Accuracy (\%),
            grid=major,
            xlabel near ticks,
            xticklabel style={/pgf/number format/1000 sep=},
            ylabel near ticks,
            xtick=data,
            yticklabel style={
                left,
                /pgf/number format/.cd,
                fixed,
                precision=2,
                /tikz/.cd
            },
            log ticks with fixed point,
            ymin=62,
            ymax=90,
          ] 
          \addplot[color=blue,mark=o] coordinates { (25, 63.1) (50,72.5) (75,77.9) (100,83.3)};
          \addplot[color=red,mark=square] coordinates { (25,65) (50,75.5) (75,78.7) (100,84.6)};
          \addplot[color=black,mark=triangle] coordinates { (25,63.6) (50,74.3) (75,81.9) (100,88.9)};                      
            \legend{Baseline, Baseline + DA, Our Model}
          \end{axis}
        \end{tikzpicture}
    }
    \caption{Color and geometric transformations}
    \label{fig_BACH_colorgeo}
\end{subfigure}
                \caption{\textbf{Classification Accuracy (\%) on BACH dataset in fonction of the amount of training data}. Our model performs better than using only predefined transformations when using the full training set. When we reduce the amount of data gradually, we can see that the amplitude of the improvement decreases for color and geometric only transformations. Below a threshold of 50\% of the training set, our model performs on par with predefined augmentations when learning color or geometric transformations but yields an inferior performance when learning both types of transformations at the same time. In this case, our model does not have enough data to learn useful transformations. }
                \label{fig:BACH_overview}
            \end{figure}

            In a first experiment, we investigate the evolution of the model accuracy with respect to the amount of training data. In Fig.\ref{fig:BACH_overview}, we can see that our model performs better than using only predefined transformations when using the full training set. When we reduce the amount of data gradually, we can see that the amplitude of the improvement decreases for color only and geometric only transformations. Below a threshold of 50\% of the training set, our model performs on par with predefined augmentations when learning color or geometric transformations only but yields an inferior performance when learning both types of transformations at the same time. In this case, our model does not have enough data to learn useful transformations. This shows that having a minimum amount of training data is a limitation and a prerequisite of our data-based learning method.

            \begin{table}[t]
                \small
                \centering
                \begin{tabular}{l|c|c}
                    BACH &  From Scratch & Pretrained model \\
                    \hline
                     Baseline & 71.60\tiny$\pm$1.29 & 83.30\tiny$\pm$1.18\\
                     Baseline + color & 75.20\tiny$\pm$1.04 & 85.10\tiny$\pm$1.19\\
                     Baseline + affine & 74.60\tiny$\pm$1.26 & 83.70\tiny$\pm$1.20\\
                     Baseline + color\&affine & 83.20\tiny$\pm$1.29 & 84.60\tiny$\pm$1.23\\
                    \hline
                     Our model (color DA) & 83.90\tiny$\pm$1.21 & 85.60\tiny$\pm$1.18\\
                     Our model (affine DA) & 82.70\tiny$\pm$1.99 & 85.40\tiny$\pm$1.25\\
                     Our model (color\&affine DA) & 85.60\tiny$\pm$1.29 & \textbf{88.90\tiny$\pm$1.25}\\
                \end{tabular}%
                \caption{\textbf{Impact of the pretraining on the classification accuracy (\%) on BACH dataset}. Transformations in parentheses are learned, others are predefined. The best classification accuracy is obtained when training a model pretrained on ImageNet. However, we can see that when training a model from scratch, the baseline accuracy is lower and using data augmentation has a bigger impact. (+14\% when training from scratch for our learned augmentations VS + 6.6\% when starting from a pretrained model.}
                \label{tab:experiments_training_from_scratch}
            \end{table}

            In a second experiment, we investigate the impact of starting from a pretrained model when training a classifier with our proposed method. In Tab.~\ref{tab:experiments_training_from_scratch}, we can see that the best classification accuracy is obtained when starting from a model pretrained on ImageNet. However, we can see that when training a model from scratch, the baseline accuracy is lower and using data augmentation has a bigger impact. (+14\% when training from scratch for our learned augmentations VS + 6.6\% when starting from a pretrained model). This experiment shows that using a pretrained model to boost the performances as usually done in the literature is helping, but using an appropriate data augmentation on top during training can further increase the final model performance.

        \subsection{Comparison with random sequences of data augmentation transformations}
            In Tab.~\ref{tab:comparison_sota}, we compare our model to a model trained with a RandAugment based framework on the six same datasets. To be fair in the comparison of the results, we limited the transformations in the RandAugment set of available transformations to the only ones that our model is learning.
            On 5 datasets, our model yields a better classification accuracy than the RandAugment based method. On Glas, both models yield similar results. Similarly to RandAugment, our model has only a few model-specific hyperparameters to tune (the augmenter network parameters). However, our model requires less prior knowledge as it does not require defining a precise list of possible transformations but works with a more generic set of differentiable transformations. Our intuition to explain the improved classification performance is that learning the optimal data augmentation for each epoch is beneficial for the model, as the time when the transformations are presented to the model is important as reported in \cite{Golatkar2019TimeMI}.\\
            
            \begin{table*}[t]
                \small
                \setlength\tabcolsep{2pt}                
                \centering
                \begin{tabular}{l|c|c|c|c|c|c}
                      & BACH & Glas & \makecell{Larynx\\ 20x} & \makecell{Larynx\\ 40x} & \makecell{Brain\\ 20x} & \makecell{Brain\\ 40x}\\
                    \hline
                    Baseline & 83.30\tiny$\pm$1.18 & 89.50\tiny$\pm$1.22 & 87.51\tiny$\pm$1.27 & 86.67\tiny$\pm$1.16 & 99.53\tiny$\pm$0.43 & 96.97\tiny$\pm$1.13\\
                    Predefined color\&affine DA & 84.60\tiny$\pm$1.23 & 98.25\tiny$\pm$1.12 & 95.24\tiny$\pm$1.25 & 95.24\tiny$\pm$1.37 & 99.53\tiny$\pm$0.43 & 98.94\tiny$\pm$1.26\\
                    \hline
                    \vspace{-4pt}
                    RandAugment & 87.25\tiny$\pm$1.48 & \textbf{99.25}\tiny$\pm$0.68 & 95.83\tiny$\pm$0.23 & 96.14\tiny$\pm$1.31 & 99.53\tiny$\pm$0.35 & 99.18\tiny$\pm$1.03\\
                    \tiny{(M,N) hyperparameters} & \tiny(3,2) & \tiny(3,2) & \tiny(4,2) & \tiny(4,2) & \tiny(3,3) & \tiny(3,3) \\
                    \hline
                    Our approach & \textbf{88.90}\tiny$\pm$1.25 & \textbf{99.25}\tiny$\pm$0.56 & \textbf{96.61}\tiny$\pm$1.23 & \textbf{97.14}\tiny$\pm$1.26 & \textbf{99.84}\tiny$\pm$0.35 & \textbf{99.24}\tiny$\pm$0.54\\
                \end{tabular}%
                \caption{\textbf{Comparison to a RandAugment based model in terms of classification accuracy (\%)}. Our model yields better results than a model trained in a RandAugment based framework on 5 datasets. On Glas it is performing on-par. Our model represents a good solution to learn the optimal data augmentation automatically for color and affine transformations.
                }
                \label{tab:comparison_sota}
            \end{table*}

    \begin{table*}
        \centering
        \begin{tabular}{cccc}
           \toprule
             & Color & Affine & Color and Affine \\
            \midrule
            BACH & \includegraphics[width=0.25\linewidth]{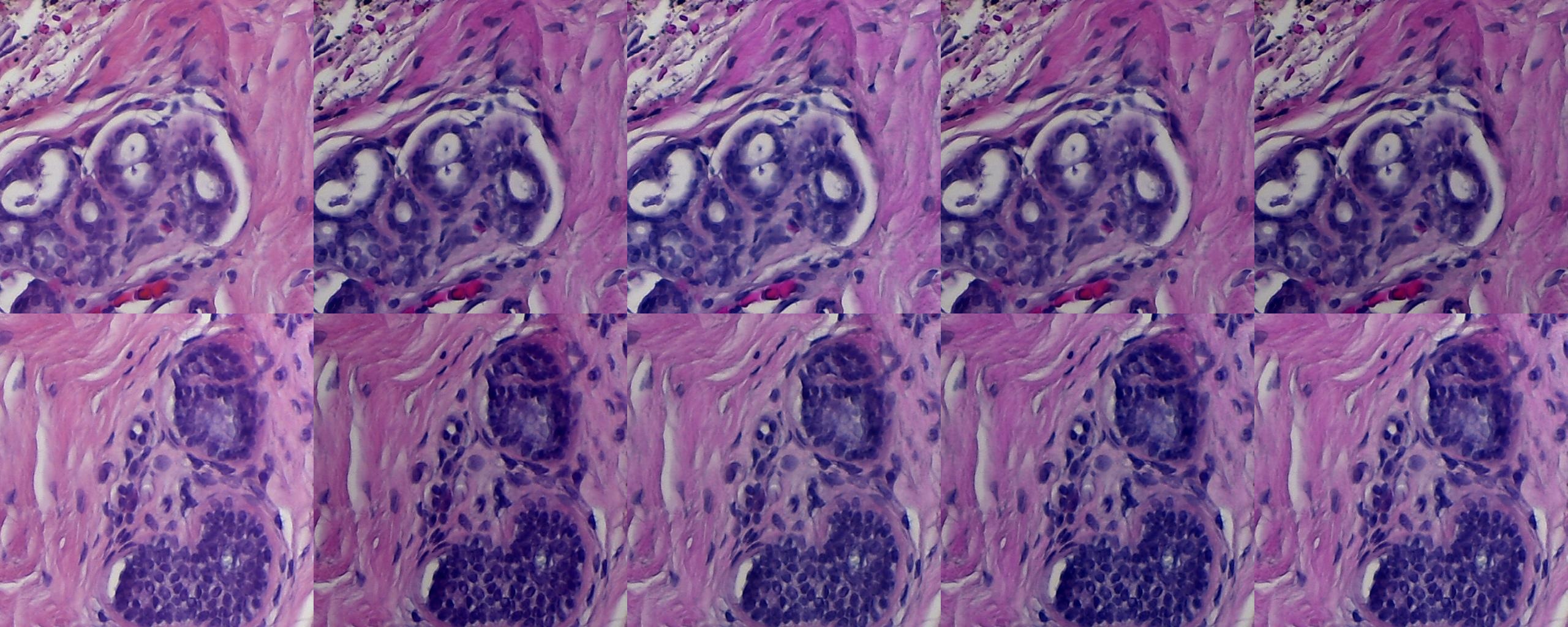} & \includegraphics[width=0.25\linewidth]{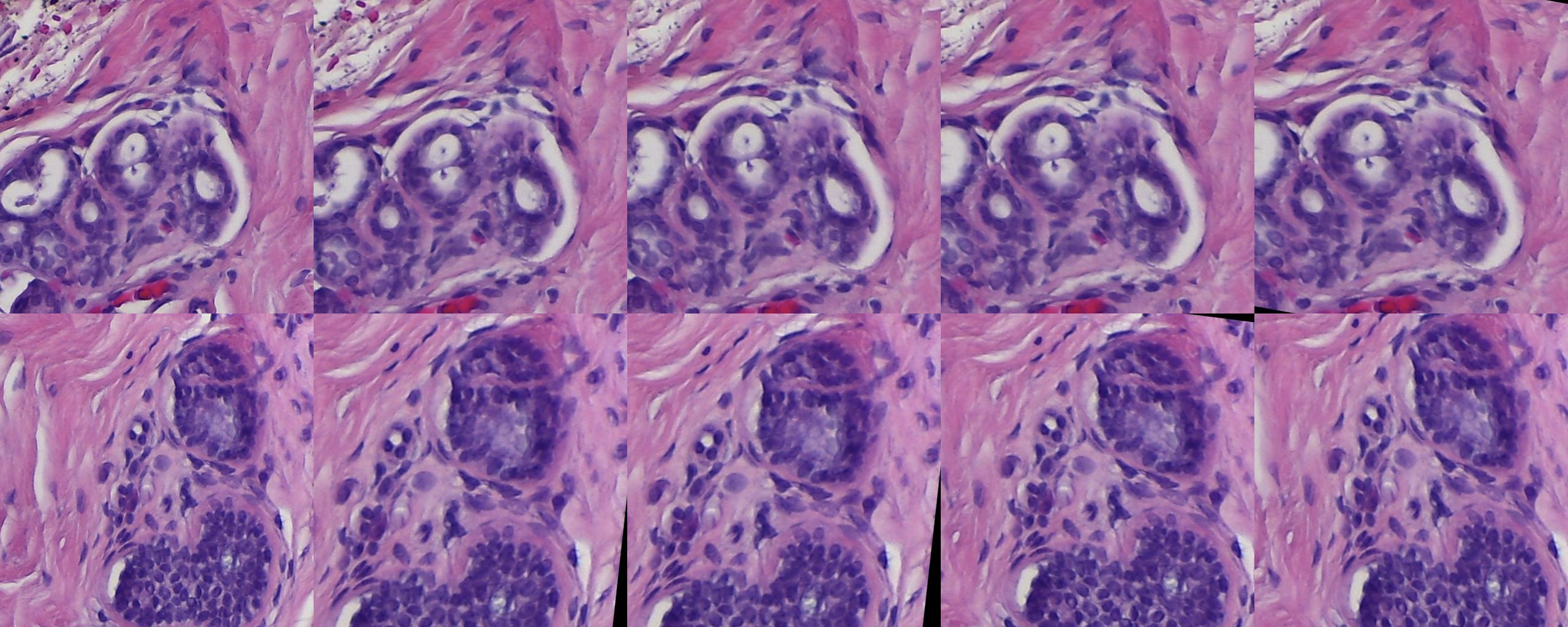} & \includegraphics[width=0.25\linewidth]{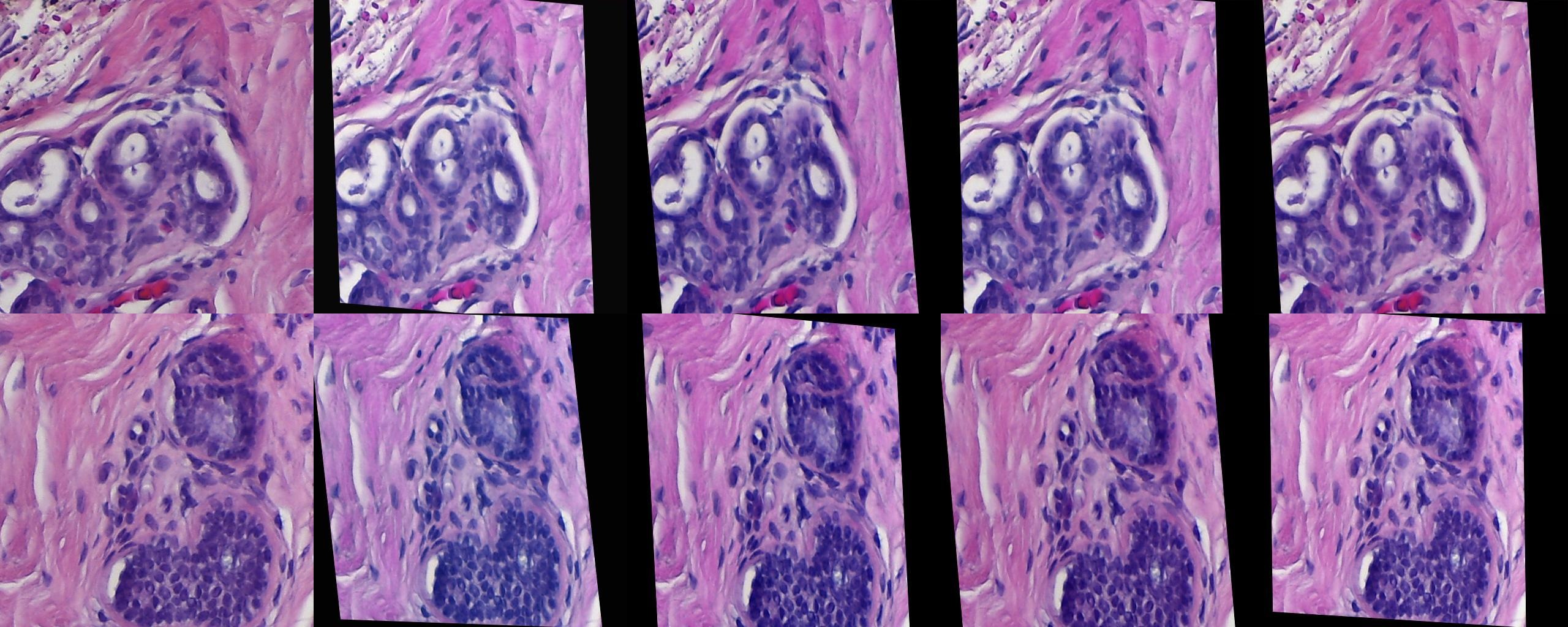} \\
            Glas & \includegraphics[width=0.25\linewidth]{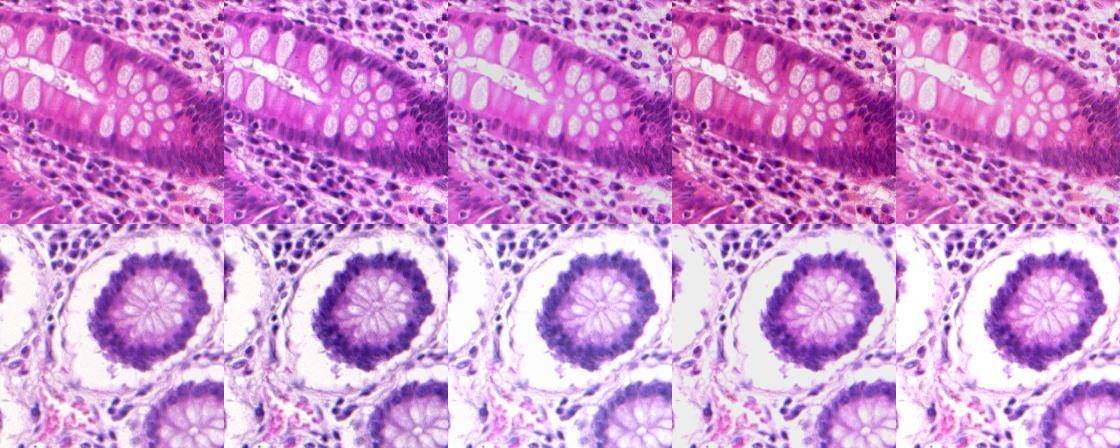} & \includegraphics[width=0.25\linewidth]{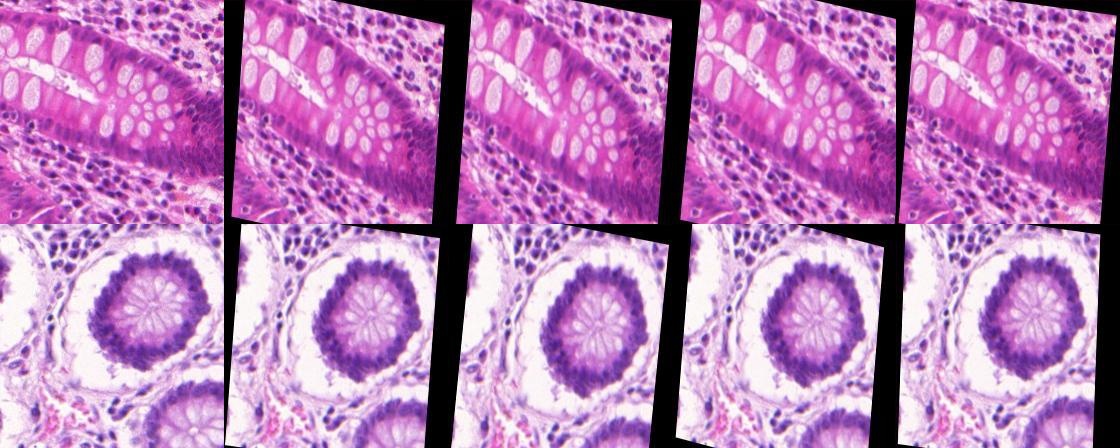} & \includegraphics[width=0.25\linewidth]{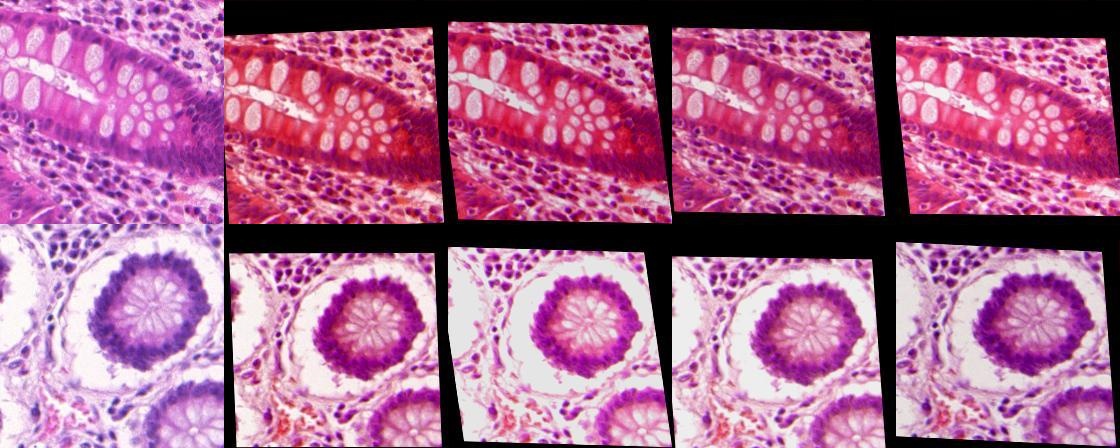} \\
            HICL Larynx20x & \includegraphics[width=0.25\linewidth]{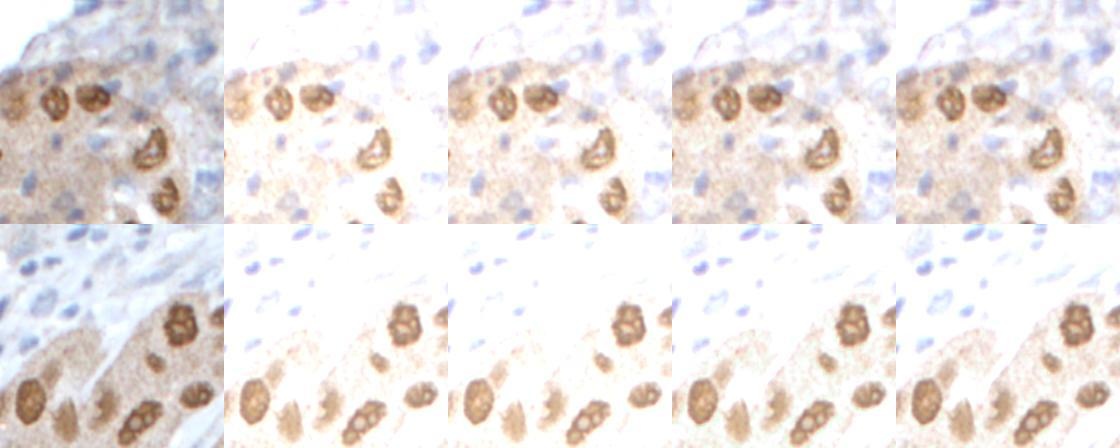} & \includegraphics[width=0.25\linewidth]{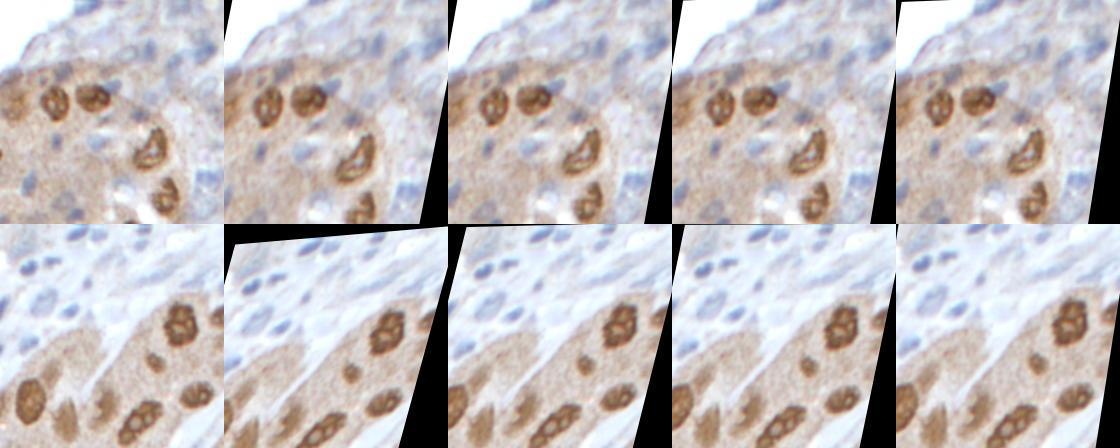} & \includegraphics[width=0.25\linewidth]{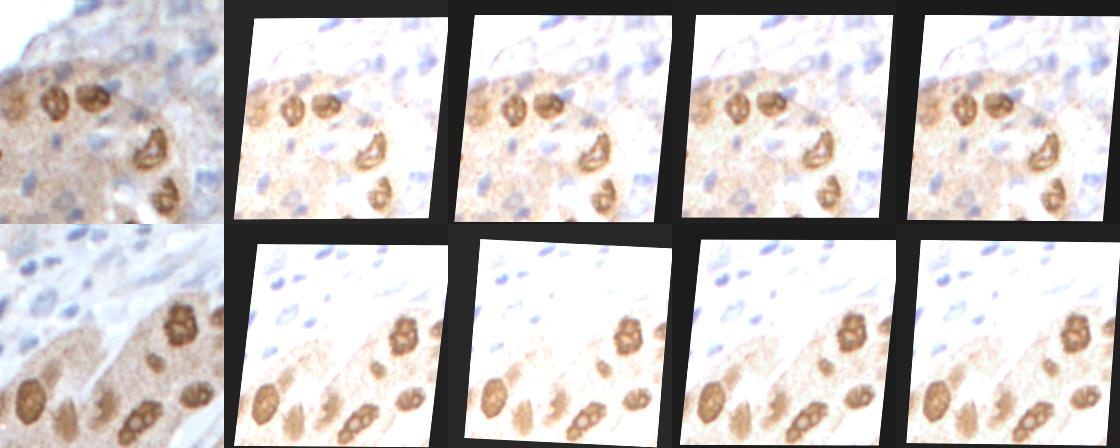} \\
            HICL Larynx40x & \includegraphics[width=0.25\linewidth]{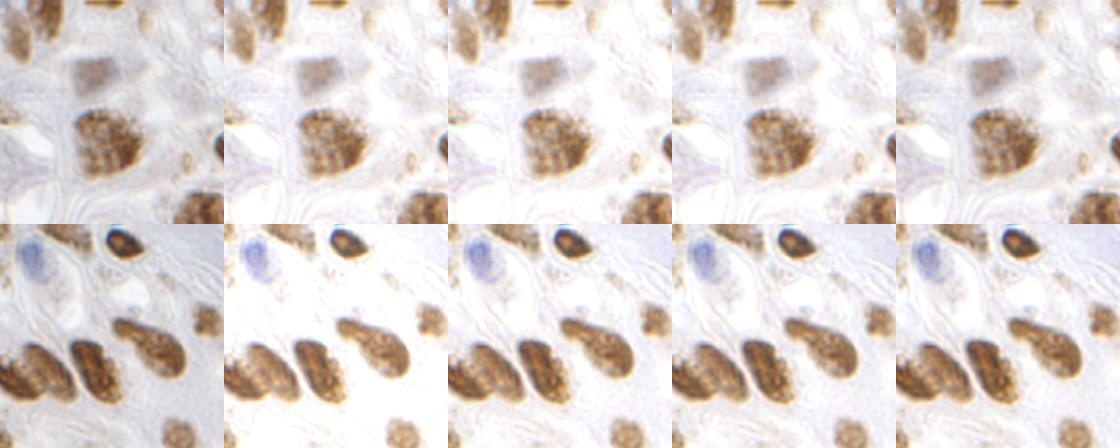} & \includegraphics[width=0.25\linewidth]{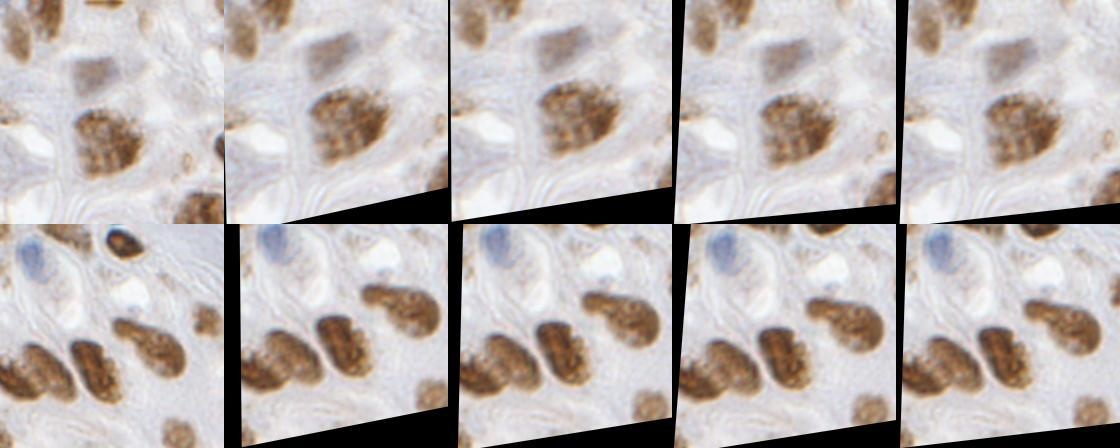} & \includegraphics[width=0.25\linewidth]{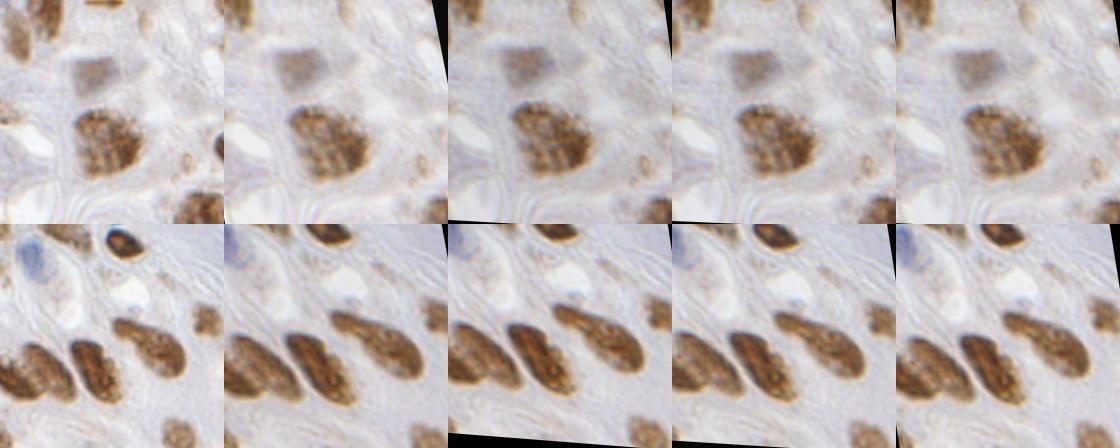} \\
            HICL Brain20x & \includegraphics[width=0.25\linewidth]{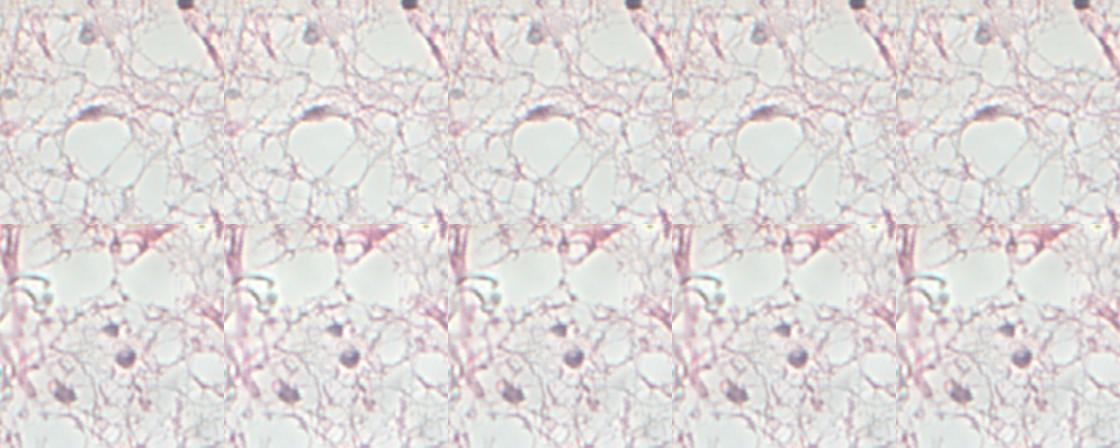} & \includegraphics[width=0.25\linewidth]{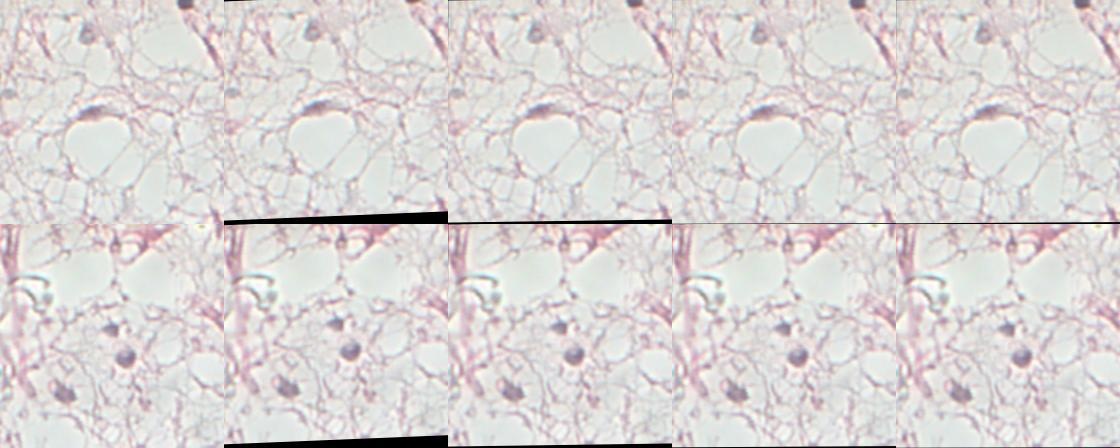} & \includegraphics[width=0.25\linewidth]{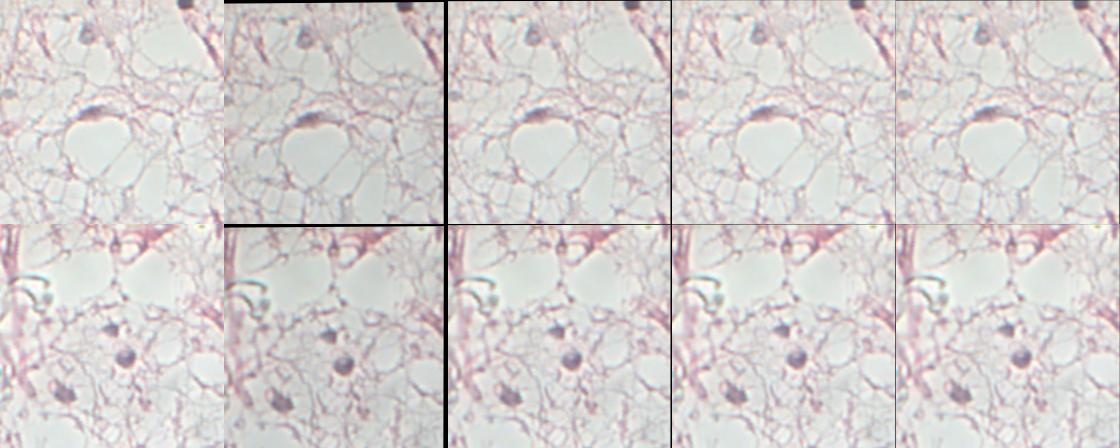} \\
            HICL Brain40x & \includegraphics[width=0.25\linewidth]{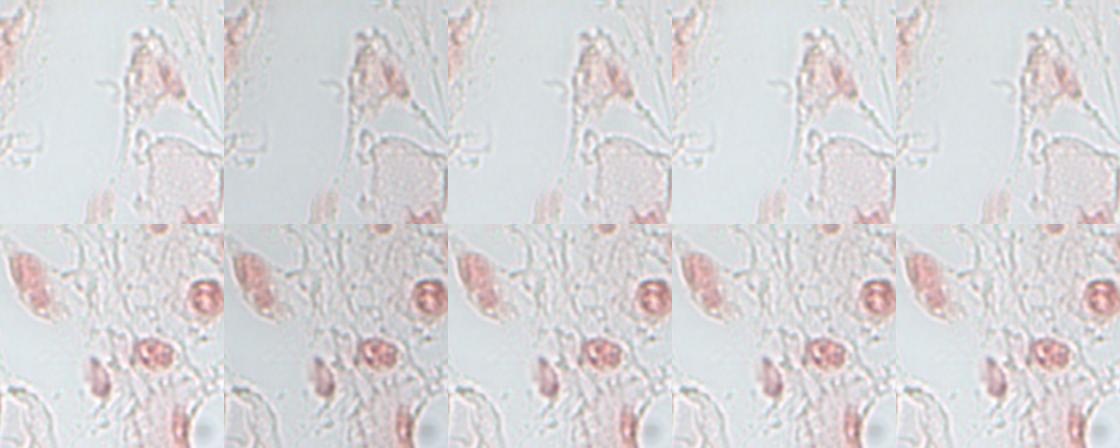} & \includegraphics[width=0.25\linewidth]{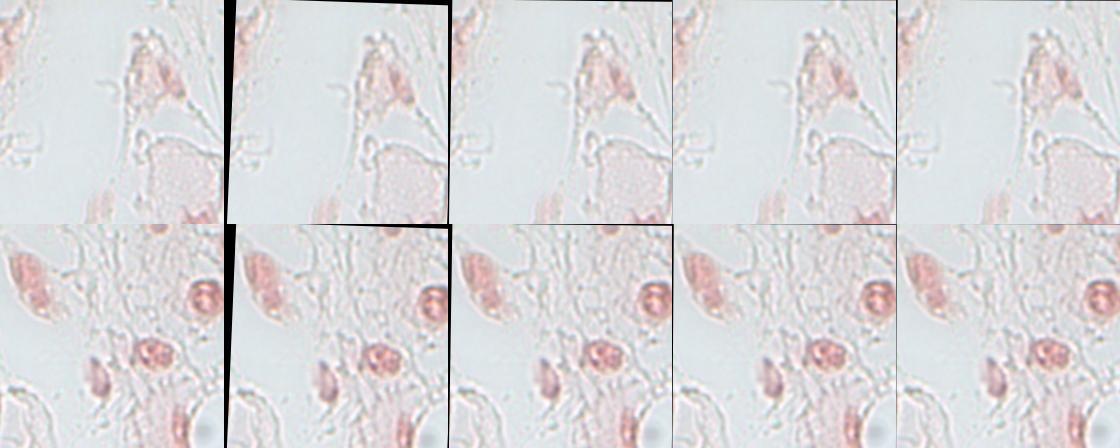} & \includegraphics[width=0.25\linewidth]{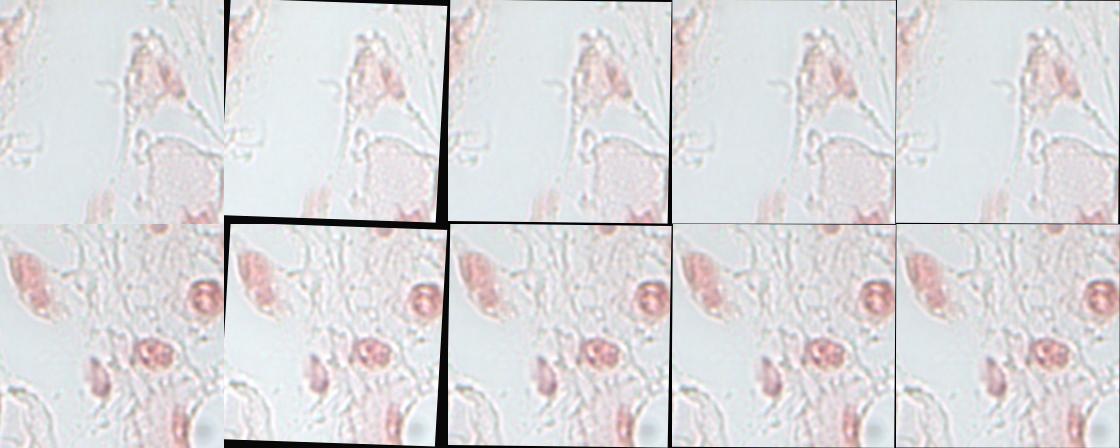} \\
            \bottomrule
        \end{tabular}
        \caption{\textbf{Qualitative results.} For each dataset and each scenario, we see the evolution of the learned transformations along the training. Transformations at the beginning of the training are stronger and tend later to finer transformations useful enough to improve the classification accuracy of the trained model. In each row, the first image is the original patch and the last one is the same patch at the end of the training. The images in-between were extracted at respectively at 25\%, 50\% and 75\% of the total number of training epochs.}
        \label{tbl:table_of_images}
    \end{table*}

    \section{Conclusion}
        We have presented a novel approach to automatically learn the transformations needed for effective data augmentation for histopathological images. The method is based on an online approximation of the bilevel optimization problem defined by alternating between optimizing the model parameters and the data augmentation hyperparameters. By doing so, we train an augmenter network to generate the right transformations at the same time as we train the classifier network. We evaluated the proposed approach on 6 different datasets with different color and affine transformations. The obtained results were comparable to or better than the results obtained with hand-defined transformations. It also yielded better results that a model trained with a RandAugment based framework. This shows that our method is very suitable in the context of histopathological images where potentially useful transformations to train a classifier are not trivial to define by hand. It also eliminates the risk to select transformations that would degrade the model accuracy.
        

\acks{This research was supported by the National Science and Engineering Research Council of Canada (NSERC), via its Discovery Grant program and MITACS via its Acc\'el\'eration program.}

%
\ethics{The work follows appropriate ethical standards in conducting research and writing the manuscript, following all applicable laws and regulations regarding treatment of animals or human subjects.}

\coi{We declare we don't have conflicts of interest.}

\bibliography{main}

\begin{thebibliography}{60}
\providecommand{\natexlab}[1]{#1}
\providecommand{\url}[1]{\texttt{#1}}
\expandafter\ifx\csname urlstyle\endcsname\relax
  \providecommand{\doi}[1]{doi: #1}\else
  \providecommand{\doi}{doi: \begingroup \urlstyle{rm}\Url}\fi

\bibitem[Antoniou et~al.(2018)Antoniou, Storkey, and
  Edwards]{antoniou2018augmenting}
Antreas Antoniou, Amos Storkey, and Harrison Edwards.
\newblock Augmenting image classifiers using data augmentation generative
  adversarial networks.
\newblock In \emph{International Conference on Artificial Neural Networks},
  2018.

\bibitem[Aresta et~al.(2019)Aresta, Ara{\'u}jo, Kwok, Chennamsetty,
  MohammedSafwanK., Varghese, Marami, Prastawa, Chan, Donovan, Fernandez,
  Zeineh, Kohl, Walz, Ludwig, Braunewell, Baust, Vu, To, Kim, Kwak, Galal,
  Sanchez-Freire, Brancati, Frucci, Riccio, Wang, Sun, Ma, Fang, Kon{\'e},
  Boulmane, Campilho, Eloy, Pol{\'o}nia, and Aguiar]{Aresta2019BACHGC}
Guilherme Aresta, Teresa Ara{\'u}jo, Scotty Kwok, Sai~Saketh Chennamsetty,
  P.~MohammedSafwanK., Alex Varghese, Bahram Marami, Marcel Prastawa, Monica
  Chan, Michael~J. Donovan, Gerardo Fernandez, Jack Zeineh, Matthias Kohl,
  Christoph Walz, Florian Ludwig, Stefan Braunewell, Maximilian Baust,
  Quoc~Dang Vu, Minh Nguyen~Nhat To, Eal Kim, Jin~Tae Kwak, Sameh Galal,
  Veronica Sanchez-Freire, Nadia Brancati, Maria Frucci, Daniel Riccio, Yaqi
  Wang, Lingling Sun, Kaiqiang Ma, Jiannan Fang, Isma{\"e}l Kon{\'e}, Lahsen
  Boulmane, Aur{\'e}lio Campilho, Catarina Eloy, Ant{\'o}nio Pol{\'o}nia, and
  Paulo Aguiar.
\newblock Bach: Grand challenge on breast cancer histology images.
\newblock \emph{Medical image analysis}, 56:\penalty0 122--139, 2019.

\bibitem[Ataky et~al.(2020)Ataky, De~Matos, Britto, Oliveira, and
  Koerich]{ataky2020data}
Steve Tsham~Mpinda Ataky, Jonathan De~Matos, Alceu de~S Britto, Luiz~ES
  Oliveira, and Alessandro~L Koerich.
\newblock Data augmentation for histopathological images based on
  gaussian-laplacian pyramid blending.
\newblock In \emph{2020 International Joint Conference on Neural Networks
  (IJCNN)}, pages 1--8. IEEE, 2020.

\bibitem[Bejnordi et~al.(2016)Bejnordi, Litjens, Timofeeva, Otte-H{\"o}ller,
  Homeyer, Karssemeijer, and van~der Laak]{Bejnordi2016StainSS}
Babak~Ehteshami Bejnordi, Geert J.~S. Litjens, Nadya Timofeeva, Irene
  Otte-H{\"o}ller, Andr{\'e} Homeyer, Nico Karssemeijer, and Jeroen van~der
  Laak.
\newblock Stain specific standardization of whole-slide histopathological
  images.
\newblock \emph{IEEE Transactions on Medical Imaging}, 35:\penalty0 404--415,
  2016.

\bibitem[Bengio(2000)]{bengio2000gradient}
Yoshua Bengio.
\newblock Gradient-based optimization of hyperparameters.
\newblock \emph{Neural computation}, 12\penalty0 (8), 2000.

\bibitem[Bergstra and Bengio(2012)]{bergstra12random}
James Bergstra and Yoshua Bengio.
\newblock Random search for hyper-parameter optimization.
\newblock \emph{J. Mach. Learn. Res.}, 13, February 2012.

\bibitem[Bergstra et~al.(2013)Bergstra, Yamins, and Cox]{bergstra2013making}
James Bergstra, Daniel Yamins, and David~Daniel Cox.
\newblock Making a science of model search: Hyperparameter optimization in
  hundreds of dimensions for vision architectures.
\newblock \emph{JMLR}, 2013.

\bibitem[Bergstra et~al.(2011)Bergstra, Bardenet, Bengio, and
  K\'{e}gl]{NIPS2011_4443}
James~S. Bergstra, R\'{e}mi Bardenet, Yoshua Bengio, and Bal\'{a}zs K\'{e}gl.
\newblock Algorithms for hyper-parameter optimization.
\newblock In \emph{Advances in Neural Information Processing Systems
  (NeurIPS)}, 2011.

\bibitem[Bertrand et~al.(2020)Bertrand, Klopfenstein, Blondel, Vaiter,
  Gramfort, and Salmon]{bertrand2020implicit}
Quentin Bertrand, Quentin Klopfenstein, Mathieu Blondel, Samuel Vaiter,
  Alexandre Gramfort, and Joseph Salmon.
\newblock Implicit differentiation of lasso-type models for hyperparameter
  optimization.
\newblock In \emph{International Conference on Machine Learning}, pages
  810--821. PMLR, 2020.

\bibitem[Chen et~al.(2020)Chen, Dobriban, and Lee]{chen2020group}
Shuxiao Chen, Edgar Dobriban, and Jane~H Lee.
\newblock A group-theoretic framework for data augmentation.
\newblock \emph{The Journal of Machine Learning Research}, 21\penalty0
  (1):\penalty0 9885--9955, 2020.

\bibitem[Chongxuan et~al.(2017)Chongxuan, Xu, Zhu, and
  Zhang]{chongxuan2017triple}
LI~Chongxuan, Taufik Xu, Jun Zhu, and Bo~Zhang.
\newblock Triple generative adversarial nets.
\newblock In \emph{Advances in Neural Information Processing Systems
  (NeurIPS)}, 2017.

\bibitem[Ciompi et~al.(2017)Ciompi, Geessink, Bejnordi, de~Souza, Baidoshvili,
  Litjens, van Ginneken, Nagtegaal, and van~der Laak]{Ciompi2017TheIO}
Francesco Ciompi, Oscar G.~F. Geessink, Babak~Ehteshami Bejnordi, Gabriel~Silva
  de~Souza, Alexi Baidoshvili, Geert J.~S. Litjens, Bram van Ginneken, Iris~D.
  Nagtegaal, and Jeroen van~der Laak.
\newblock The importance of stain normalization in colorectal tissue
  classification with convolutional networks.
\newblock \emph{2017 IEEE 14th International Symposium on Biomedical Imaging
  (ISBI 2017)}, pages 160--163, 2017.

\bibitem[Colson et~al.(2007)Colson, Marcotte, and Savard]{colson2007overview}
Beno{\^\i}t Colson, Patrice Marcotte, and Gilles Savard.
\newblock An overview of bilevel optimization.
\newblock \emph{Annals of operations research}, 153\penalty0 (1), 2007.

\bibitem[Cubuk et~al.(2019{\natexlab{a}})Cubuk, Zoph, Mane, Vasudevan, and
  Le]{Cubuk_2019_CVPR}
Ekin~D. Cubuk, Barret Zoph, Dandelion Mane, Vijay Vasudevan, and Quoc~V. Le.
\newblock Autoaugment: Learning augmentation strategies from data.
\newblock In \emph{Computer Vision and Pattern Recognition (CVPR)},
  2019{\natexlab{a}}.

\bibitem[Cubuk et~al.(2019{\natexlab{b}})Cubuk, Zoph, Shlens, and
  Le]{Cubuk2019RandaugmentPA}
Ekin~Dogus Cubuk, Barret Zoph, Jonathon Shlens, and Quoc~V. Le.
\newblock Randaugment: Practical automated data augmentation with a reduced
  search space.
\newblock \emph{2020 IEEE/CVF Conference on Computer Vision and Pattern
  Recognition Workshops (CVPRW)}, pages 3008--3017, 2019{\natexlab{b}}.

\bibitem[de~Matos et~al.(2021)de~Matos, Ataky, de~Souza~Britto, Soares~de
  Oliveira, and Lameiras~Koerich]{electronics10050562}
Jonathan de~Matos, Steve Tsham~Mpinda Ataky, Alceu de~Souza~Britto,
  Luiz~Eduardo Soares~de Oliveira, and Alessandro Lameiras~Koerich.
\newblock Machine learning methods for histopathological image analysis: A
  review.
\newblock \emph{Electronics}, 10\penalty0 (5), 2021.
\newblock ISSN 2079-9292.

\bibitem[Domke(2012)]{domke2012generic}
Justin Domke.
\newblock Generic methods for optimization-based modeling.
\newblock In \emph{Artificial Intelligence and Statistics}, 2012.

\bibitem[Faryna et~al.(2021)Faryna, van~der Laak, and
  Litjens]{faryna2021tailoring}
Khrystyna Faryna, Jeroen van~der Laak, and Geert Litjens.
\newblock Tailoring automated data augmentation to h\&e-stained histopathology.
\newblock In \emph{Medical Imaging with Deep Learning}, 2021.

\bibitem[Finn et~al.(2017)Finn, Abbeel, and Levine]{finn2017model}
Chelsea Finn, Pieter Abbeel, and Sergey Levine.
\newblock Model-agnostic meta-learning for fast adaptation of deep networks.
\newblock In \emph{International Conference on Machine Learning (ICML)}, 2017.

\bibitem[Franceschi et~al.(2017)Franceschi, Donini, Frasconi, and
  Pontil]{pmlr-v70-franceschi17a}
Luca Franceschi, Michele Donini, Paolo Frasconi, and Massimiliano Pontil.
\newblock Forward and reverse gradient-based hyperparameter optimization.
\newblock In \emph{International Conference on Machine Learning (ICML)}, 2017.

\bibitem[Franceschi et~al.(2018)Franceschi, Frasconi, Salzo, Grazzi, and
  Pontil]{pmlr-v80-franceschi18a}
Luca Franceschi, Paolo Frasconi, Saverio Salzo, Riccardo Grazzi, and
  Massimiliano Pontil.
\newblock Bilevel programming for hyperparameter optimization and
  meta-learning.
\newblock In \emph{International Conference on Machine Learning (ICML)}, 2018.

\bibitem[Frid-Adar et~al.(2018)Frid-Adar, Klang, Amitai, Goldberger, and
  Greenspan]{FridAdar2018SyntheticDA}
Maayan Frid-Adar, Eyal Klang, Michal~Marianne Amitai, Jacob Goldberger, and
  Hayit Greenspan.
\newblock Synthetic data augmentation using gan for improved liver lesion
  classification.
\newblock \emph{2018 IEEE 15th International Symposium on Biomedical Imaging
  (ISBI 2018)}, pages 289--293, 2018.

\bibitem[Garcea et~al.(2022)Garcea, Serra, Lamberti, and
  Morra]{GARCEA2022106391}
Fabio Garcea, Alessio Serra, Fabrizio Lamberti, and Lia Morra.
\newblock Data augmentation for medical imaging: A systematic literature
  review.
\newblock \emph{Computers in Biology and Medicine}, page 106391, 2022.

\bibitem[Glotsos et~al.(2008)Glotsos, Kalatzis, Spyridonos, Kostopoulos,
  Daskalakis, Athanasiadis, Ravazoula, Nikiforidis, and
  Cavouras]{Glotsos2008ImprovingAI}
Dimitris~T. Glotsos, Ioannis Kalatzis, Panagiota Spyridonos, Spiros
  Kostopoulos, Antonis Daskalakis, Emmanouil~I. Athanasiadis, Panagiota
  Ravazoula, George Nikiforidis, and Dionisis~A. Cavouras.
\newblock Improving accuracy in astrocytomas grading by integrating a robust
  least squares mapping driven support vector machine classifier into a two
  level grade classification scheme.
\newblock \emph{Computer methods and programs in biomedicine}, 90 3:\penalty0
  251--61, 2008.

\bibitem[Golatkar et~al.(2019)Golatkar, Achille, and
  Soatto]{Golatkar2019TimeMI}
Aditya Golatkar, Alessandro Achille, and Stefano Soatto.
\newblock Time matters in regularizing deep networks: Weight decay and data
  augmentation affect early learning dynamics, matter little near convergence.
\newblock In \emph{Advances in Neural Information Processing Systems
  (NeurIPS)}, 2019.

\bibitem[Goodfellow et~al.(2014)Goodfellow, Pouget-Abadie, Mirza, Xu,
  Warde-Farley, Ozair, Courville, and Bengio]{Goodfellow2014GenerativeAN}
Ian~J. Goodfellow, Jean Pouget-Abadie, Mehdi Mirza, Bing Xu, David
  Warde-Farley, Sherjil Ozair, Aaron~C. Courville, and Yoshua Bengio.
\newblock Generative adversarial nets.
\newblock In \emph{Advances in Neural Information Processing Systems
  (NeurIPS)}, 2014.

\bibitem[Hataya et~al.(2019)Hataya, Zdenek, Yoshizoe, and
  Nakayama]{Hataya2019FasterAL}
Ryuichiro Hataya, Jan Zdenek, Kazuki Yoshizoe, and Hideki Nakayama.
\newblock Faster autoaugment: Learning augmentation strategies using
  backpropagation.
\newblock In \emph{European Conference on Computer Vision (ECCV)}, 2019.

\bibitem[Hataya et~al.(2022)Hataya, Zdenek, Yoshizoe, and
  Nakayama]{Hataya_2022_WACV}
Ryuichiro Hataya, Jan Zdenek, Kazuki Yoshizoe, and Hideki Nakayama.
\newblock Meta approach to data augmentation optimization.
\newblock In \emph{Proceedings of the IEEE/CVF Winter Conference on
  Applications of Computer Vision (WACV)}, pages 2574--2583, January 2022.

\bibitem[He et~al.(2015)He, Zhang, Ren, and Sun]{He2015ResNet}
Kaiming He, Xiangyu Zhang, Shaoqing Ren, and Jian Sun.
\newblock Deep residual learning for image recognition.
\newblock \emph{Computer Vision and Pattern Recognition (CVPR)}, 2015.

\bibitem[Ho et~al.(2019)Ho, Liang, Stoica, Abbeel, and Chen]{ho2019pba}
Daniel Ho, Eric Liang, Ion Stoica, Pieter Abbeel, and Xi~Chen.
\newblock Population based augmentation: Efficient learning of augmentation
  policy schedules.
\newblock In \emph{International Conference on Machine Learning (ICML)}, 2019.

\bibitem[Hutter et~al.(2011)Hutter, Hoos, and
  Leyton-Brown]{hutter2011sequential}
Frank Hutter, Holger~H Hoos, and Kevin Leyton-Brown.
\newblock Sequential model-based optimization for general algorithm
  configuration.
\newblock In \emph{International Conference on Learning and Intelligent
  Optimization}, 2011.

\bibitem[Jaderberg et~al.(2015)Jaderberg, Simonyan, Zisserman,
  et~al.]{jaderberg2015spatial}
Max Jaderberg, Karen Simonyan, Andrew Zisserman, et~al.
\newblock Spatial transformer networks.
\newblock In \emph{Advances in Neural Information Processing Systems
  (NeurIPS)}, 2015.

\bibitem[Ker et~al.(2018)Ker, Wang, Rao, and Lim]{ker2018deep}
Justin Ker, Lipo Wang, Jai Rao, and Tchoyoson Lim.
\newblock Deep learning applications in medical image analysis.
\newblock \emph{IEEE Access}, 6:\penalty0 9375--9389, 2018.

\bibitem[Lim et~al.(2019)Lim, Kim, Kim, Kim, and Kim]{lim2019fast}
Sungbin Lim, Ildoo Kim, Taesup Kim, Chiheon Kim, and Sungwoong Kim.
\newblock Fast autoaugment.
\newblock In \emph{Advances in Neural Information Processing Systems
  (NeurIPS)}, 2019.

\bibitem[Lin et~al.(2019)Lin, Guo, Li, Wu, Lin, Ouyang, and
  Yan]{Lin2019OnlineHL}
Chen Lin, Minghao Guo, Chuming Li, Wei Wu, Dahua Lin, Wanli Ouyang, and Junjie
  Yan.
\newblock Online hyper-parameter learning for auto-augmentation strategy.
\newblock \emph{2019 IEEE/CVF International Conference on Computer Vision
  (ICCV)}, pages 6578--6587, 2019.

\bibitem[Litjens et~al.(2017)Litjens, Kooi, Bejnordi, Setio, Ciompi,
  Ghafoorian, van~der Laak, van Ginneken, and S{\'{a}}nchez]{Litjens2017}
Geert J.~S. Litjens, Thijs Kooi, Babak~Ehteshami Bejnordi, Arnaud
  Arindra~Adiyoso Setio, Francesco Ciompi, Mohsen Ghafoorian, Jeroen A. W.~M.
  van~der Laak, Bram van Ginneken, and Clara~I. S{\'{a}}nchez.
\newblock A survey on deep learning in medical image analysis.
\newblock \emph{Medical Image Analysis}, 42:\penalty0 60--88, 2017.
\newblock \doi{10.1016/j.media.2017.07.005}.
\newblock URL \url{https://doi.org/10.1016/j.media.2017.07.005}.

\bibitem[Liu et~al.(2019)Liu, Simonyan, and Yang]{liu2018darts}
Hanxiao Liu, Karen Simonyan, and Yiming Yang.
\newblock Darts: Differentiable architecture search.
\newblock In \emph{International Conference on Learning Representations
  (ICLR)}, 2019.

\bibitem[Luketina et~al.(2016)Luketina, Berglund, Greff, and
  Raiko]{luketina2016scalable}
Jelena Luketina, Mathias Berglund, Klaus Greff, and Tapani Raiko.
\newblock Scalable gradient-based tuning of continuous regularization
  hyperparameters.
\newblock In \emph{International Conference on Machine Learning (ICML)}, pages
  2952--2960, 2016.

\bibitem[MacKay et~al.(2019)MacKay, Vicol, Lorraine, Duvenaud, and
  Grosse]{mackay2019self}
Matthew MacKay, Paul Vicol, Jon Lorraine, David Duvenaud, and Roger Grosse.
\newblock Self-tuning networks: Bilevel optimization of hyperparameters using
  structured best-response functions.
\newblock In \emph{International Conference on Learning Representations
  (ICLR)}, 2019.

\bibitem[Maclaurin et~al.(2015)Maclaurin, Duvenaud, and
  Adams]{pmlr-v37-maclaurin15}
Dougal Maclaurin, David Duvenaud, and Ryan Adams.
\newblock Gradient-based hyperparameter optimization through reversible
  learning.
\newblock In \emph{International Conference on Machine Learning (ICML)}, 2015.

\bibitem[Mirza and Osindero(2014)]{mirza2014conditional}
Mehdi Mirza and Simon Osindero.
\newblock Conditional generative adversarial nets.
\newblock \emph{arXiv preprint arXiv:1411.1784}, 2014.

\bibitem[Mounsaveng et~al.(2021)Mounsaveng, Laradji, Ayed, V{\'a}zquez, and
  Pedersoli]{Mounsaveng2021LearningDA}
Saypraseuth Mounsaveng, Issam~H. Laradji, Ismail~Ben Ayed, David V{\'a}zquez,
  and Marco Pedersoli.
\newblock Learning data augmentation with online bilevel optimization for image
  classification.
\newblock \emph{2021 IEEE Winter Conference on Applications of Computer Vision
  (WACV)}, pages 1690--1699, 2021.

\bibitem[Ninos et~al.(2015)Ninos, Kostopoulos, Kalatzis, Sidiropoulos,
  Ravazoula, Sakellaropoulos, Panayiotakis, Economou, and
  Cavouras]{Ninos2015MicroscopyIA}
Konstantinos Ninos, Spiros Kostopoulos, Ioannis Kalatzis, Konstantinos
  Sidiropoulos, Panagiota Ravazoula, George Sakellaropoulos, George~S.
  Panayiotakis, George Economou, and Dionisis~A. Cavouras.
\newblock Microscopy image analysis of p63 immunohistochemically stained
  laryngeal cancer lesions for predicting patient 5-year survival.
\newblock \emph{European Archives of Oto-Rhino-Laryngology}, 273:\penalty0
  159--168, 2015.

\bibitem[Odena et~al.(2017)Odena, Olah, and Shlens]{DBLP:conf/icml/OdenaOS17}
Augustus Odena, Christopher Olah, and Jonathon Shlens.
\newblock Conditional image synthesis with auxiliary classifier gans.
\newblock In \emph{International Conference on Machine Learning (ICML)}, 2017.

\bibitem[Paszke et~al.(2019)Paszke, Gross, Massa, Lerer, Bradbury, Chanan,
  Killeen, Lin, Gimelshein, Antiga, Desmaison, Kopf, Yang, DeVito, Raison,
  Tejani, Chilamkurthy, Steiner, Fang, Bai, and Chintala]{NEURIPS2019_9015}
Adam Paszke, Sam Gross, Francisco Massa, Adam Lerer, James Bradbury, Gregory
  Chanan, Trevor Killeen, Zeming Lin, Natalia Gimelshein, Luca Antiga, Alban
  Desmaison, Andreas Kopf, Edward Yang, Zachary DeVito, Martin Raison, Alykhan
  Tejani, Sasank Chilamkurthy, Benoit Steiner, Lu~Fang, Junjie Bai, and Soumith
  Chintala.
\newblock Pytorch: An imperative style, high-performance deep learning library.
\newblock In \emph{Advances in Neural Information Processing Systems 32}, pages
  8024--8035. Curran Associates, Inc., 2019.
\newblock URL
  \url{http://papers.neurips.cc/paper/9015-pytorch-an-imperative-style-high-performance-deep-learning-library.pdf}.

\bibitem[Pathak et~al.(2015)Pathak, Kr{\"a}henb{\"u}hl, and
  Darrell]{Pathak2015ConstrainedCN}
Deepak Pathak, Philipp Kr{\"a}henb{\"u}hl, and Trevor Darrell.
\newblock Constrained convolutional neural networks for weakly supervised
  segmentation.
\newblock \emph{2015 IEEE International Conference on Computer Vision (ICCV)},
  pages 1796--1804, 2015.

\bibitem[Pedregosa(2016)]{pmlr-v48-pedregosa16}
Fabian Pedregosa.
\newblock Hyperparameter optimization with approximate gradient.
\newblock In \emph{International Conference on Machine Learning (ICML)}, 2016.

\bibitem[Ratner et~al.(2017)Ratner, Ehrenberg, Hussain, Dunnmon, and
  R{\'e}]{ratner2017learning}
Alexander~J Ratner, Henry Ehrenberg, Zeshan Hussain, Jared Dunnmon, and
  Christopher R{\'e}.
\newblock Learning to compose domain-specific transformations for data
  augmentation.
\newblock In \emph{Advances in Neural Information Processing Systems
  (NeurIPS)}, 2017.

\bibitem[Riba et~al.(2019)Riba, Mishkin, Ponsa, Rublee, and
  Bradski]{Riba2019KorniaAO}
Edgar Riba, Dmytro Mishkin, Daniel Ponsa, Ethan Rublee, and Gary~R. Bradski.
\newblock Kornia: an open source differentiable computer vision library for
  pytorch.
\newblock \emph{2020 IEEE Winter Conference on Applications of Computer Vision
  (WACV)}, pages 3663--3672, 2019.

\bibitem[Rony et~al.(2023)Rony, Belharbi, Dolz, Ayed, McCaffrey, and
  Granger]{Rony2022DeepWL}
J{\'e}r{\^o}me Rony, Soufiane Belharbi, Jos{\'e} Dolz, Ismail~Ben Ayed, Luke
  McCaffrey, and Eric Granger.
\newblock Deep weakly-supervised learning methods for classiﬁcation and
  localization in histology images: A comparative study.
\newblock In \emph{Medical Imaging with Deep Learning (MIDL)}, 2023.

\bibitem[Shaban et~al.(2019)Shaban, Cheng, Hatch, and
  Boots.]{Shaban-AISTATS-19}
Amirreza Shaban, Ching-An Cheng, Nathan Hatch, and Byron Boots.
\newblock Truncated backpropogation for bi-level optimization.
\newblock In \emph{Proceedings of the 22nd International Conference on
  Artificial Intelligence and Statistics}, 2019.

\bibitem[Shao et~al.(2022)Shao, Montasser, and Blum]{Shao2022ATO}
Hang Shao, Omar Montasser, and Avrim Blum.
\newblock A theory of pac learnability under transformation invariances.
\newblock \emph{ArXiv}, abs/2202.07552, 2022.

\bibitem[Shen et~al.(2017)Shen, Wu, and Suk]{shen2017deep}
Dinggang Shen, Guorong Wu, and Heung-Il Suk.
\newblock Deep learning in medical image analysis.
\newblock \emph{Annual review of biomedical engineering}, 19:\penalty0
  221--248, 2017.

\bibitem[Sirinukunwattana et~al.(2016)Sirinukunwattana, Pluim, Chen, Qi, Heng,
  Guo, Wang, Matuszewski, Bruni, Sanchez, B{\"o}hm, Ronneberger, Cheikh,
  Racoceanu, Kainz, Pfeiffer, Urschler, Snead, and
  Rajpoot]{Sirinukunwattana2016GlandSI}
Korsuk Sirinukunwattana, Josien P.~W. Pluim, Hao Chen, Xiaojuan Qi, Pheng-Ann
  Heng, Yun~Bo Guo, Li~Yang Wang, Bogdan~J. Matuszewski, Elia Bruni, Urko
  Sanchez, Anton B{\"o}hm, Olaf Ronneberger, Bassem~Ben Cheikh, Daniel
  Racoceanu, Philipp Kainz, Michael Pfeiffer, Martin Urschler, David R.~J.
  Snead, and Nasir~M. Rajpoot.
\newblock Gland segmentation in colon histology images: The glas challenge
  contest.
\newblock \emph{Medical Image Analysis}, 35:\penalty0 489–502, 2016.

\bibitem[Snoek et~al.(2012)Snoek, Larochelle, and Adams]{snoek12practical}
Jasper Snoek, Hugo Larochelle, and Ryan~P Adams.
\newblock Practical bayesian optimization of machine learning algorithms.
\newblock In \emph{Advances in Neural Information Processing Systems
  (NeurIPS)}, 2012.

\bibitem[Szeliski(2010)]{Szeliski2010ComputerV}
Richard Szeliski.
\newblock Computer vision - algorithms and applications.
\newblock In \emph{Texts in Computer Science}, 2010.

\bibitem[Tellez et~al.(2019)Tellez, Litjens, B{\'a}ndi, Bulten, Bokhorst,
  Ciompi, and van~der Laak]{Tellez2019QuantifyingTE}
David Tellez, Geert J.~S. Litjens, P{\'e}ter B{\'a}ndi, Wouter Bulten, J.~M.
  Bokhorst, Francesco Ciompi, and Jeroen van~der Laak.
\newblock Quantifying the effects of data augmentation and stain color
  normalization in convolutional neural networks for computational pathology.
\newblock \emph{Medical image analysis}, 58:\penalty0 101544, 2019.

\bibitem[Tran et~al.(2017)Tran, Pham, Carneiro, Palmer, and
  Reid]{tran2017bayesian}
Toan Tran, Trung Pham, Gustavo Carneiro, Lyle Palmer, and Ian Reid.
\newblock A bayesian data augmentation approach for learning deep models.
\newblock In \emph{Advances in Neural Information Processing Systems
  (NeurIPS)}, 2017.

\bibitem[Wagner et~al.(2021)Wagner, Khalili, Sharma, Boxberg, Marr, Back, and
  Peng]{wagner2021structure}
Sophia~J Wagner, Nadieh Khalili, Raghav Sharma, Melanie Boxberg, Carsten Marr,
  Walter~de Back, and Tingying Peng.
\newblock Structure-preserving multi-domain stain color augmentation using
  style-transfer with disentangled representations.
\newblock In \emph{International Conference on Medical Image Computing and
  Computer-Assisted Intervention}, pages 257--266. Springer, 2021.

\bibitem[Williams and Peng(1990)]{Williams90anefficient}
Ronald~J. Williams and Jing Peng.
\newblock An efficient gradient-based algorithm for on-line training of
  recurrent network trajectories.
\newblock \emph{Neural Computation}, 1990.

\end{thebibliography}


\clearpage
\appendix

\end{document}